\documentclass[runningheads]{llncs}
\usepackage{graphicx}
\usepackage{comment}
\usepackage{amsmath,amssymb} 
\usepackage{color}

\usepackage{epsfig}
\usepackage{graphicx}
\usepackage{booktabs}
\usepackage{lipsum}
\usepackage{multirow}
\usepackage{footnote}
\usepackage{mathtools}
\usepackage{algorithm} 
\usepackage{algpseudocode} 
\usepackage{subfig}
\usepackage{float}
\usepackage{adjustbox}
\usepackage{import}
\usepackage{caption}
\usepackage{url}            

\newfloat{algorithm}{t}{lop}

\DeclareMathOperator{\EX}{\mathbb{E}}

\DeclareMathOperator*{\argmin}{arg\,min}
\DeclareMathOperator{\prox}{\text{Prox}}
\DeclareMathOperator{\sign}{sign}

\DeclareMathOperator{\diag}{diag}

\usepackage{amssymb}
\usepackage{pifont}
\newcommand{\cmark}{\ding{51}}%
\newcommand{\xmark}{\ding{55}}%

\newcommand\vsp{\vspace*{-0.0cm}}
\newcommand\yb{\mathbf{y}}
\newcommand\Yb{\mathbf{Y}}
\newcommand\xb{\mathbf{x}}
\newcommand\Db{\mathbf{D}}
\newcommand\Ab{\mathbf{A}}
\newcommand\Ub{\mathbf{U}}
\newcommand\Zb{\mathbf{Z}}
\newcommand\ub{\mathbf{u}}

\newcommand\zb{\mathbf{z}}
\newcommand\Cb{\mathbf{C}}

\newcommand\Wb{\mathbf{W}}
\newcommand\Rb{\mathbf{R}}
\newcommand\Mb{\mathbf{M}}
\newcommand\Bb{\mathbf{B}}
\newcommand\db{\mathbf{d}}

\newcommand\ones{\mathbf{1}}
\newcommand\Real{\mathbb{R}}
\newcommand\alphab{\boldsymbol{\alpha}}
\newcommand\betab{\boldsymbol{\beta}}
\newcommand\Lambdab{\boldsymbol{\Lambda}}
\newcommand\Thetab{\boldsymbol{\Theta}}
\newcommand\Sigmab{\boldsymbol{\Sigma}}
\newcommand\kappab{\boldsymbol{\kappa}}

\newcommand\Pcal{{\mathcal{P}}}

\begin{document}
\pagestyle{headings}
\mainmatter
\def\ECCVSubNumber{4019}  

\title{Fully Trainable and Interpretable Non-Local Sparse Models for Image  Restoration} 

\titlerunning{Trainable Non-Local Sparse Models for Image  Restoration}

\author{Bruno Lecouat \inst{1,2} \and
Jean Ponce \inst{1} \and
Julien Mairal\inst{2}}

\authorrunning{B. Lecouat, J. Ponce, and J. Mairal}

\institute{Inria, \'Ecole normale sup\'erieure, CNRS, PSL University, 75005 Paris, France
\and
Inria, Univ. Grenoble Alpes, CNRS, Grenoble INP, LJK, 38000 Grenoble, France
}

\maketitle

\begin{abstract}
Non-local self-similarity and sparsity principles have proven to be 
powerful priors for natural image modeling. We propose a novel 
differentiable relaxation of joint sparsity that exploits both 
principles and leads to a general framework for image restoration which 
is (1) trainable end to end, (2) fully interpretable, and (3) much more 
compact than competing deep learning architectures. We apply this 
approach to denoising, blind denoising, jpeg deblocking, and demosaicking, and show that, 
with as few as 100K parameters, its performance on several standard 
benchmarks is on par or better than state-of-the-art methods that may 
have an order of magnitude or more parameters.

\end{abstract} 
 
\section{Introduction}
\begin{figure*}[t]
\def \width{2.3}
\def \2{14}
\def \1{23}
\def \2{14}
\def \3{07}
\def \4{20}

\def \width{2.0}
\def \Width{2.0}

\centering
\def \width{2.3}
\def \2{14}
\def \1{23}
\def \2{14}
\def \3{07}
\def \4{20}
\def \width{2.3}
\def \Width{2.3}

\centering
\begin{tabular}{ccccccc}
\includegraphics[width=\width cm]{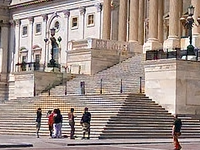} & 
\includegraphics[width=\width cm]{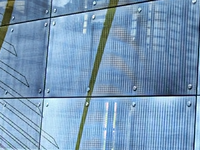} &
\includegraphics[width=\Width cm]{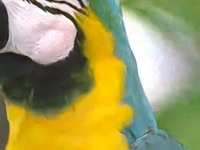}&
\includegraphics[width=\Width cm]{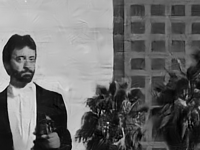}&
\includegraphics[width=\width cm]{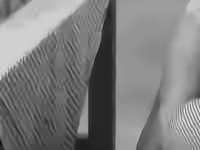} &

\\
\includegraphics[width=\width cm]{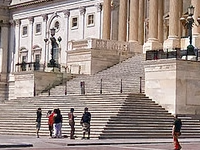} & 
\includegraphics[width=\width cm]{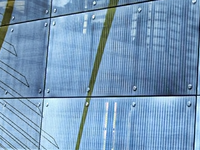}&
\includegraphics[width=\Width cm]{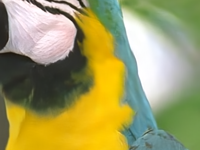} &
\includegraphics[width=\Width cm]{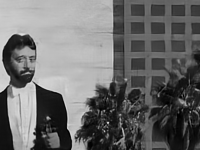}&
\includegraphics[width=\width cm]{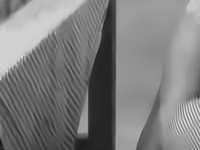} & 

 \\

\multicolumn{2}{c}{Demosaicking} & \multicolumn{2}{c}{Denoising} & Jpeg deblocking   \\

\end{tabular}
\caption{Effect of combining sparse and non-local priors for different reconstruction tasks. Top: reconstructions with sparse prior only, exhibiting artefacts. Bottom: reconstruction with both priors, artefact-free. Best seen in color by zooming on a computer screen.}
\label{fig:mosaic}
\vspace*{-0.2cm}
\end{figure*}

The image processing community has long focused on designing
 handcrafted models of natural images to address inverse problems,
leading, for instance, to  
differential operators~\cite{perona1990}, total
variation~\cite{rudin1992nonlinear}, or wavelet sparsity~\cite{mallat} approaches. More recently,
image restoration para\-digms have shifted towards data-driven
approaches. For instance, non-local
means~\cite{buades2005non} exploits self-similarities, 
and many successful approaches have relied on
unsupervised methods such as learned sparse models~\cite{aharon2006k,mairal2014sparse},
Gaussian scale mixtures~\cite{portilla}, or fields of experts~\cite{roth}.
More powerful models such as BM3D~\cite{dabov2009bm3d} have also been obtained by combining several priors, in particular self-similarities and sparse
representations 
\cite{dabov2007image,dabov2009bm3d,dong2012nonlocally,gu2014weighted,mairal2009non}.

These methods are now often outperformed by
deep learning models, which are able to leverage pairs of corrupted/clean
images for supervised learning,
in  tasks such as
denoising~\cite{lefkimmiatis2017non,liu2018non,plotz2018neural,zhang2017beyond},
demoisaicking~\cite{kokkinos2019iterative,zhang2017learning,zhang2019rnan},
upsampling~\cite{dong2015image,kim2016accurate}, or artefact removal~\cite{zhang2019rnan}.  Yet, they also suffer from lack of
interpretability and the need to learn a huge number of parameters.
Improving these two aspects is
one of the key motivation of this paper.  Our goal is to design algorithms that bridge the gap in performance between earlier approaches
that are parameter-efficient and interpretable, and current deep
models. 

Specifically, we propose a differentiable relaxation of the non-local sparse
model LSSC~\cite{mairal2009non}. 
The relaxation allows us to obtain models that may be trained end-to-end,
and which admit a simple interpretation in terms of joint sparse coding
of similar patches.
The principle of end-to-end training for sparse coding was
introduced in~\cite{mairal2011task}, and later combined in~\cite{wang2015deep} for super-resolution
with variants of the LISTA
algorithm~\cite{chen2018theoretical,gregor2010learning,liu2018alista}.
A variant based on convolutional sparse coding was then proposed
in~\cite{simon2019rethinking} for image denoising, and another one based on
the K-SVD algorithm~\cite{elad2006image} was introduced in~\cite{scetbon2019deep}.
Note that these works are part of a vast litterature on model-inspired methods,
 where the model architecture is related to an optimization strategy for minimizing an objective,
  see~\cite{lefkimmiatis2017non,sun2016deep,venkatakrishnan2013plug}. 

In contrast, our main contribution is to extend the idea of differentiable
algorithms to \emph{structured} sparse models~\cite{jenatton2011structured},
which is a key concept behind the LSSC, CSR, and BM3D approaches.
To the best of our knowledge, this is the first time that non-local sparse
models are shown to be effective in a supervised learning setting.
As~\cite{scetbon2019deep}, we argue that bridging classical successful image
priors within deep learning frameworks is a key to overcome the limitations of
current state-of-the-art models.  A striking fact is notably the performance of 
the resulting models given their low number of parameters. 

For example, our method for image denoising performs on par with the
deep learning baseline DnCNN~\cite{zhang2017beyond} with $8$x less
parameters, significantly outperforms the color variant CDnCNN with $6$x
less parameters, and achieves state-of-the-art results for blind
denoising and jpeg deblocking. 
For these two last tasks, relying on an interpretable model is important;
most parameters are devoted to image reconstruction and can be shared 
by models dedicated to different noise levels. Only a small subset of parameters can be
seen as regularization parameters, and may be made noise-dependent,
thus removing the burden of training several large independent models for each noise level.
For image demosaicking, we obtain similar results
as the state-of-the-art approach RNAN~\cite{zhang2019rnan}, while reducing the
number of parameters by $76$x.
Perhaps more important than improving the PSNR, the
principle of non local sparsity also reduces visual artefacts when compared to
using sparsity alone,
which is illustrated in Figure~\ref{fig:mosaic}.

Our models are implemented in PyTorch and our code can be found at \url{ https://github.com/bruno-31/groupsc}.

\section{Preliminaries and Related Work}\label{sec:related}

In this section, we introduce non-local sparse coding models for image denoising and present a differentiable algorithm for sparse coding~\cite{gregor2010learning}.

\vsp
\paragraph{Sparse coding models on learned dictionaries.}

A simple approach for image denoising introduced in~\cite{elad2006image} consists of assuming that
natural image patches can be well approximated by linear combinations
of few dictionary elements. Thus, a clean estimate of a noisy patch is obtained by computing a sparse approximation.
Given a noisy image, we denote by $\yb_1,\ldots,\yb_n$ the set of~$n$
overlapping patches of size $\sqrt{m} \times \sqrt{m}$, which we represent
by vectors in~$\Real^m$ for grayscale images. Each patch is then processed by
solving the sparse decomposition problem
\begin{equation}\label{eq:l1_problem}
   \min_{\alphab_i \in \Real^{p}} \frac{1}{2} \|\yb_i-\Db\alphab_i\|_2^2+\lambda\|\alphab_i\|_1,
\end{equation}
where $\Db=[\db_1,\ldots,\db_p]$ in $\Real^{m \times p}$ is the dictionary,
which we assume given at the moment, and $\|.\|_1$ is the $\ell_1$-norm, which is known to encourage
sparsity, see~\cite{mairal2014sparse}.
Note that a direct sparsity measure such as $\ell_0$-penalty may also be used,
at the cost of producing a combinatorially hard problem, whereas~(\ref{eq:l1_problem}) is convex.

Then, $\Db\alphab_i$ is a clean estimate of $\yb_i$. Since the patches overlap, we obtain
$m$ estimates for each pixel and the denoised image is obtained by averaging:
\begin{equation}
   \hat{\xb} = \frac{1}{m} \sum_{i=1}^n \Rb_i \Db \alphab_i, \label{eq:averaging}
\end{equation}
where $\Rb_i$ is a linear operator that places the patch $\Db \alphab_i$ at the position centered on pixel $i$ on the image.
Note that for simplicity, we neglect the fact that pixels close to the
image border admit less estimates, unless zero-padding is used.

Whereas we have previously assumed that a good dictionary~$\Db$ for natural images is
available, the authors of~\cite{elad2006image} have proposed
to learn $\Db$  by solving a matrix factorization
problem called \emph{dictionary learning}~\cite{field}.

\vsp
\paragraph{Differentiable algorithms for sparse coding.}
ISTA~\cite{figueiredo2003} is
a popular algorithm to solve problem~(\ref{eq:l1_problem}),
which alternates between gradient descent steps with respect to the smooth term of~(\ref{eq:l1_problem})
and the soft-thresholding operator $S_\eta(x)=  \sign(x) \max(0,|x| - \eta)$.

Note that such a step performs an affine transformation followed
by the pointwise non-linear function $S_{\eta}$, which makes it tempting to
consider $K$ steps of the algorithm, see it as a neural network with $K$
layers, and learn the corresponding weights.
Following such an insight, the authors of~\cite{gregor2010learning} have proposed the LISTA algorithm,
which is trained such that the resulting neural network learns to approximate the
solution of~(\ref{eq:l1_problem}).
Other variants were then proposed, see~\cite{chen2018theoretical,liu2018alista}; as~\cite{simon2019rethinking}, the one we have adopted may be written as
\begin{equation}
   \label{eq:recurrent}
   \alphab_i^{(k+1)} = S_{\Lambdab_{k}} \left [\alphab_i^{(k)}+ \Cb^\top\left(\yb_i-\Db\alphab_i^{(k)}\right) \right ],
\end{equation}
where $\Cb$ has the same size as $\Db$ and $\Lambdab_{k}$ in $\Real^p$ is such that
$S_{\Lambdab_{k}}$ performs a soft-thresholding operation with a
different threshold for each vector entry. Then,
the variables $\Cb, \Db$ and $\Lambdab_{k}$ are learned for a supervised image reconstruction task.

Note that when $\Cb = \eta\Db$ and $\Lambdab_{k}=\eta\lambda \ones$, where $\eta$ is a step size,
the recursion
recovers exactly the ISTA algorithm. Empirically, it has been observed that allowing
$\Cb \neq \Db$ accelerates convergence and could be
interpreted as learning a pre-conditioner for ISTA~\cite{liu2018alista}, whereas allowing
$\Lambdab_{k}$ to have entries different than~$\lambda \eta$ corresponds to using a weighted $\ell_1$-norm
and learning the weights.

There have been already a few attempts to leverage the LISTA algorithm  for specific  image restoration tasks such as
super-resolution~\cite{wang2015deep} or denoising~\cite{simon2019rethinking}, which we
extend in our paper with non-local priors and structured sparsity.

\vsp
\paragraph{Exploiting self-similarities.}
The non-local means approach \cite{buades2005non} consists of averaging similar patches
that are corrupted by i.i.d. zero-mean noise,
such that averaging reduces the noise variance without corrupting the signal.
The intuition relies on the fact that natural images
admit many local self-similarities. This is a non-parametric
approach (technically a Nadaraya-Watson estimator), which can be used to reduce the number of parameters of deep learning models.

\begin{figure*}[t]
   \def \width{5.8}
   \centering
   \includegraphics[width=\width cm]{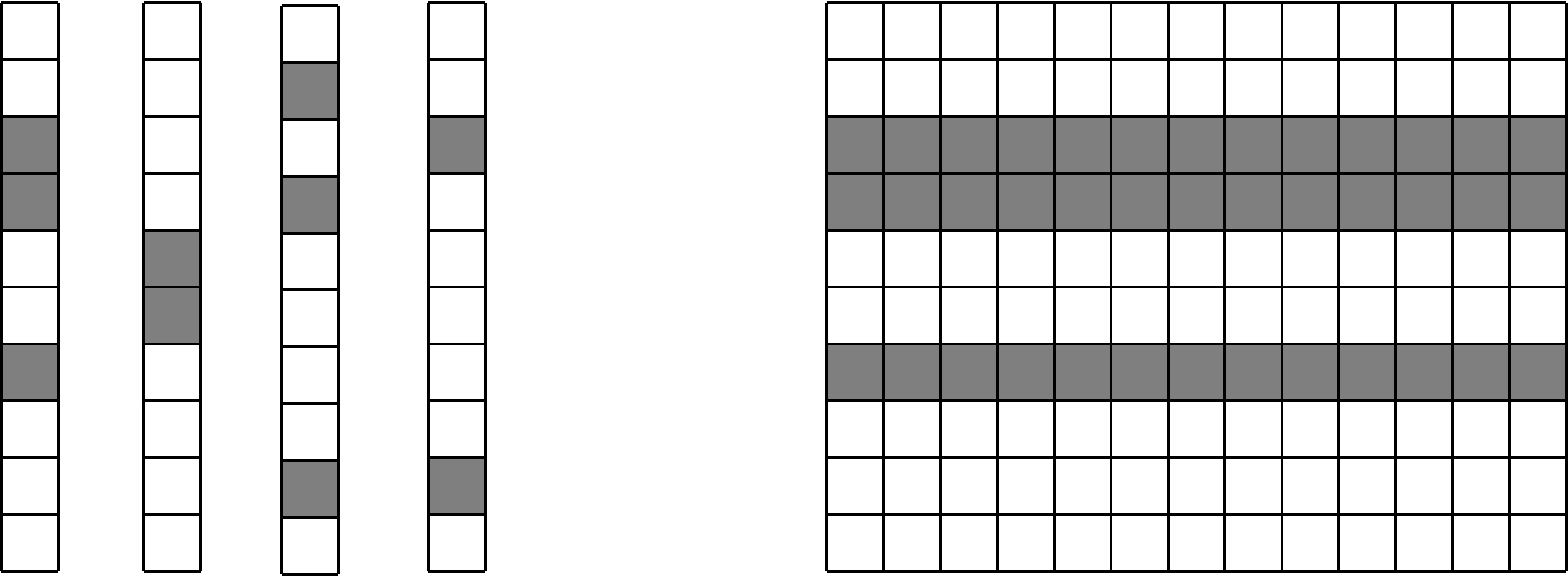}
   \label{fig:demos}
   \caption{(Left) sparsity pattern of codes with grey values representing non-zero entries; (right) group sparsity of codes for similar patches. Figure from \cite{mairal2009non}. }
   \label{fig:sparse}
\end{figure*}

\vsp
\paragraph{Non local sparse models.}
The LSSC approach~\cite{mairal2009non} relies on the principle of
joint sparsity.
Denoting by $S_i$ a set of patches similar to $\yb_i$ according to some criterion,

we consider the matrix
$\Ab_{i}=[\alphab_l]_{l \in S_i}$  in $\Real^{p\times |S_i|}$ of corresponding coefficients.
LSSC encourages the codes
$\{\alphab_l\}_{l \in S_i}$ to share the same sparsity pattern---that is, the set of non-zero entries.
This can be achieved by using a group-sparsity regularizer
\begin{equation}
   \label{groupeq}
   \|\Ab_i\|_{1,2} =   \sum_{j=1}^p \|\Ab_i^j\|_2,
\end{equation}
where $\Ab_i^j$ is the $j$-th row in $\Ab_i$.
The effect of this norm is to encourage sparsity patterns
to be shared across similar patches, as illustrated in Figure~\ref{fig:sparse}.
It may be seen as a convex relaxation of the number of non-zero rows in~$\Ab_i$, see~\cite{mairal2009non}.

Building a differentiable algorithm relying on both sparsity and non-local self-similarities
is challenging, as the clustering approach used by LSSC (or CSR) is typically not a continuous operation
of the dictionary parameters.

\vsp
\paragraph{Deep learning models.}

In the context of image restoration, successful principles for deep learning models include very deep networks, batch norm, and residual learning~\cite{lefkimmiatis2018universal,zhang2017beyond,zhang2018ffdnet,zhang2019rnan}.  Recent models also use attention mechanisms to model self similarities, which are pooling operations akin to non-local means.
More precisely, a non local
module has been proposed in~\cite{liu2018non}, which performs weighed average
of similar features, and in \cite{plotz2018neural}, a relaxation
of the k-nearest selection rule is introduced for similar purposes.

\vsp
\paragraph{Model-based methods.}
Unfolding an optimization algorithm to design an inference architecture is not limited to sparse coding.
For instance \cite{sun2016deep,zhang2018ista} propose trainable architectures based on unrolled ADMM.
The authors of \cite{lefkimmiatis2017non,lefkimmiatis2018universal} propose a deep learning architecture
inspired from proximal gradient descent in order to solve a constrained optimization problem for denoising;
\cite{chen2016trainable} optimize hyperparameters of non linear reaction diffusion models; \cite{bertocchi2019deep}
unroll an interior point algorithm. Finally,  Plug-and-Play \cite{venkatakrishnan2013plug} is a framework for
image restoration exploiting a denoising prior as a modular part of model-based optimization methods to solve
various inverse problems. Several works leverage the plug-in principle with half quadratic spliting
\cite{zoran2011learning}, deep denoisers \cite{zhang2017learning}, message passing algorithms
\cite{fletcher2018plug}, or augmented Lagrangian \cite{romano2017little}.

\section{Proposed Approach}

We now present trainable sparse coding models for image denoising,
following~\cite{simon2019rethinking}, with a few minor improvements,
before introducing differentiable relaxations for the LSSC method \cite{mairal2009non} .
%
A different approach to take into account self similarities in sparse models is
the CSR approach \cite{dong2012nonlocally}. We have
empirically observed that it does not perform as well as LSSC. Nevertheless, we believe it to be
conceptually interesting, and provide a brief description in the appendix.

\subsection{Trainable Sparse Coding (without Self-Similarities)}\label{subsec:sc}
In \cite{simon2019rethinking}, the sparse coding approach (SC) is combined with the LISTA algorithm to perform
denoising tasks.\footnote{Specifically, \cite{simon2019rethinking} proposes a model based on
   convolutional sparse coding (CSC).
   CSC is a variant of SC, where a full image is
   approximated by a linear combination of small dictionary elements. Unfortunately, CSC leads to
   ill-conditioned optimization problems and has shown to perform poorly
   for image denoising.  For this reason,~\cite{simon2019rethinking} introduces
   a hybrid approach between SC and CSC.
   In our paper, we have decided to use the SC baseline and leave the investigation of CSC models for
   future work.}
The only modification we introduce here is a centering step
for the patches, which empirically yields better results.

\begin{figure}[t]
   \centering
   \includegraphics[width=\linewidth]{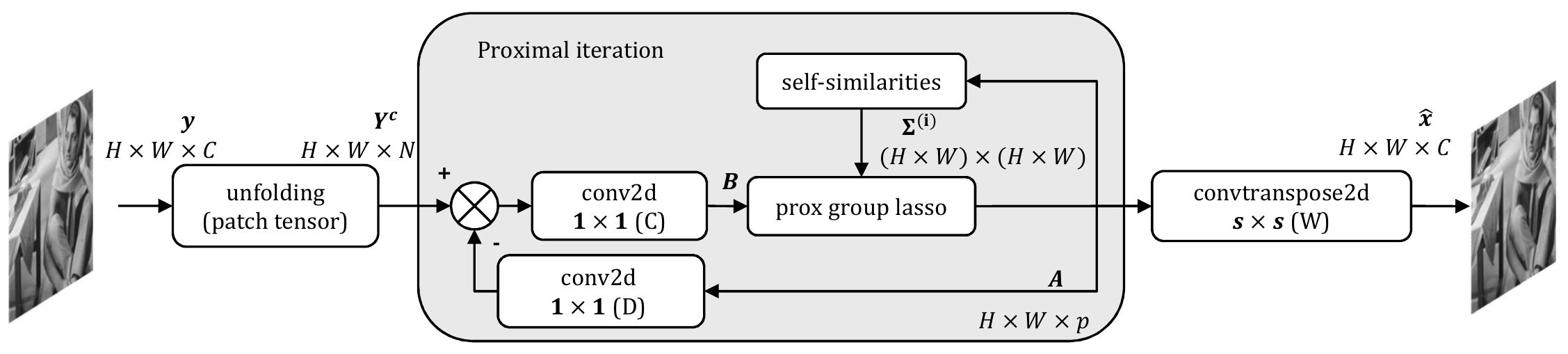}
   \caption{An illustration of the main inference algorithm for GroupSC.
      See Figure \ref{fig:nlmodule} for an illustration of the self-similarity module.}\label{fig:diagram}
\end{figure}

\begin{figure}[t] 
   \centering
   \includegraphics[width=0.80\linewidth]{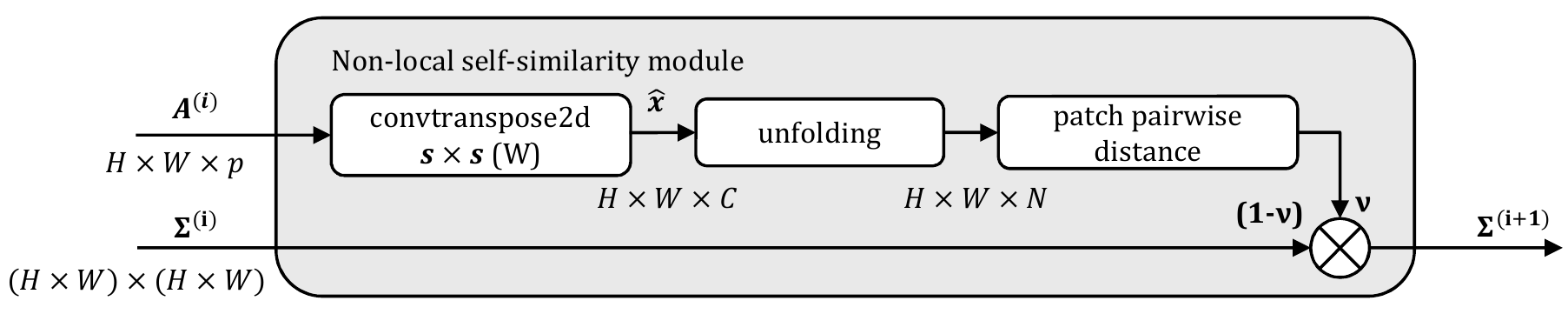}
   \caption{An illustration of the self-similarity module used in our GroupSC algorithm.}\label{fig:nlmodule}
\end{figure}

\vsp
\paragraph{SC Model - inference with fixed parameters.}
Following the approach and notation from Section~\ref{sec:related},
the first step consists of extracting all overlapping patches $\yb_1,\ldots,\yb_n$. 
Then, we perform the centering operation for every patch
\begin{equation}
   \yb_i^{c} \triangleq \yb_i - \mu_i \ones_m ~~~\text{with}~~~~ \mu_i \triangleq \frac{1}{m} \ones_m^\top  \yb_i.\label{eq:centering}
\end{equation}
The mean value~$\mu_i$ is recorded and added back after denoising $\yb_i^c$. Hence,
low-frequency components do not flow through the model.
The centering step is not used in~\cite{simon2019rethinking}, but we have
found it to be useful. 

The next step consists of sparsely encoding each centered patch $\yb_i^c$ with
$K$ steps of the LISTA variant presented in~(\ref{eq:recurrent}), replacing $\yb_i$
by $\yb_i^c$ there, assuming the parameters $\Db, \Cb$ and $\Lambdab_{k}$ are
given.
Here, a minor change compared to~\cite{simon2019rethinking}
is the use of varying parameters $\Lambdab_{k}$ at each LISTA step. 
Finally, the final image is obtained by averaging the patch estimates as in~(\ref{eq:averaging}),
after adding back $\mu_i$:
\begin{equation}
   \hat{\xb} = \frac{1}{n} \sum_{i=1}^N \Rb_i (\Wb \alphab_i^{(K)} + \mu_i \ones_m), \label{eq:averaging2}
\end{equation}
but the dictionary~$\Db$ is replaced by another matrix~$\Wb$.
The reason for decoupling $\Db$ from $\Wb$ is that the $\ell_1$ penalty used by the LISTA method is known to shrink the coefficients~$\alphab_i$
too much.
For this reason, classical
denoising approaches such as~\cite{elad2006image,mairal2009non} use instead the $\ell_0$-penalty,
but we have found it ineffective for
end-to-end training. Therefore, as in~\cite{simon2019rethinking},
we have chosen
to decouple $\Wb$ from~$\Db$.

\vsp
\paragraph{Training the parameters.}

We now assume that we are given a training set of pairs of
clean/noisy images $(\xb,\yb) \sim \Pcal$, and we minimize in a supervised fashion
\begin{equation}
   \min _{\Thetab}  \EX_{(\xb,\yb) \sim \Pcal}  \left\| \hat{\xb}(\yb) - \xb\right\|_2^2,\label{eq:cost}
\end{equation}
where $\Thetab =
   \{\Cb,\Db,\Wb,(\Lambdab_{k})_{k=0,1 \dots K-1}, \kappab, \nu \}$ is the set of parameters to learn and $\hat{\xb}$ is the denoised image defined in~(\ref{eq:averaging2}).

\subsection{Differentiable Relaxation for Non-Local Sparse Priors}
\begin{algorithm}
   \caption{Pseudo code for the inference model of GroupSC.}\label{alg:pseudo}
   \begin{algorithmic}[1]
      \State Extract patches $\Yb=[\yb_1,\ldots,\yb_n]$ and center them with~(\ref{eq:centering});
      \State Initialize the codes $\alphab_i$ to $0$;
      \State Initialize image estimate $\hat{\xb}$ to the noisy input $\yb$;
      \State Initialize pairwise similarities~$\Sigmab$ between patches of $\hat{\xb}$;
      \For {$k=1,2,\ldots K$}
      \State Compute pairwise patch similarities~$\hat{\Sigmab}$ on~$\hat{\xb}$;
      \State Update $\Sigmab \leftarrow (1-\nu) \Sigmab + \nu \hat{\Sigmab}$;
      \For {$i=1,2,\ldots,N$ in parallel}
      \State $\alphab_i \leftarrow \prox_{\Sigmab, \Lambdab_k} \left [\alphab_i+ \Cb^\top(\yb_i^c-\Db\alphab_i) \right]$;
      \EndFor
      \State Update the denoised image~$\hat{\xb}$ by averaging~(\ref{eq:averaging2});
      \EndFor
   \end{algorithmic}
\end{algorithm}

Self-similarities are modeled by replacing the $\ell_1$-norm by
structured sparsity-inducing regularization functions. In Algorithm~\ref{alg:pseudo}, we
present a generic approach to use this principle within a supervised learning approach,
based on a similarity matrix $\Sigmab$,
overcoming the difficulty of hard clustering/grouping patches together. In Figure~\ref{fig:diagram}, we also provide a diagram of one step of the inference algorithm.
At each step, the method computes pairwise patch similarities~$\Sigmab$
between patches of a current estimate~$\hat{\xb}$, using various possible
metrics that we discuss in Section~\ref{sec:similarities}.
The codes~$\alphab_i$ are updated by computing a so-called proximal operator, defined below, for a particular
penalty that depends on $\Sigmab$ and some parameters $\Lambdab_k$. Practical variants
where the pairwise similarities are only updated once in a while, are discussed in Section~\ref{sec:extensions}.
\begin{definition}[Proximal operator]
   Given a convex function~$\Psi: \Real^p \!\to\! \Real$, the proximal operator of $\Psi$ is defined~as the unique solution of
   \begin{equation}
      \prox_{\Psi}[\zb] = \argmin_{\ub \in \Real^p} \frac{1}{2}\|\zb-\ub\|^2 + \Psi(\ub).
   \end{equation}
\end{definition}
The proximal operator plays a key role in optimization and admits a closed form
for many penalties, see~\cite{mairal2014sparse}.
Indeed, given~$\Psi$, it may be shown that the iterations
$\alphab_i \leftarrow \prox_{\eta\Psi} \left [\alphab_i+ \eta \Db^\top(\yb_i^c-\Db\alphab_i) \right]$
are instances of the ISTA algorithm~\cite{beck2009fast} for minimizing
$$ \min_{\alphab_i \in \Real^p} \frac{1}{2}\|\yb_i^c - \Db\alphab_i\|^2 + \Psi(\alphab_i),$$
and the update of $\alphab_i$ in Algorithm~\ref{alg:pseudo} simply extend LISTA to deal with~$\Psi$.
Note that for the weighted $\ell_1$-norm $\Psi(\ub) = \sum_{j=1}^p \lambda_j \left | \ub[j]  \right | $, the proximal operator is the soft-thresholding
operator $S_{\Lambda}$ introduced in Section~\ref{sec:related} for $\Lambdab = (\lambda_1,\ldots,\lambda_p)$ in $\Real^p$,
and we simply recover the SC algorithm from Section~\ref{subsec:sc} since $\Psi$ does not depend on the pairwise similarities~$\Sigmab$. Next, we present different structured sparsity-inducing penalties that yield more  effective algorithms.

\vsp
\subsubsection{Group-SC.}\label{subsec:groupsc}
For each location~$i$, the LSSC approach~\cite{mairal2009non} defines
groups of similar patches
$S_i \triangleq \left \{ j=1, \dots, n \: \text{s.t.} \:  \|\yb_i-\yb_j||_2^2 \leq \xi \right \}$ for some threshold $\xi$.
For computational reasons,
LSSC relaxes this definition in practice, and implements a clustering method
such that $S_i=S_j$ if $i$ and $j$ belong to the same group.
Then, under this clustering assumption and given a dictionary~$\Db$, LSSC minimizes
\begin{equation} \label{optim_pblm}
   \min_{\Ab} \frac{1}{2} \|\Yb^c-\Db \Ab\|_{\text{F}}^2 + \sum_{i=1}^N \Psi_i(\Ab) ~~\text{with}~~ \Psi_i(\Ab)\!=\!\lambda_i\|\Ab_i\|_{1,2},
\end{equation}
where $\Ab \!=\![\alphab_1,\ldots,\alphab_N]$ in $\Real^{m \times N}$ represents all codes, $\Ab_i\!=\![\alphab_l]_{l \in S_i}$,
$\|.\|_{1,2}$ is the group sparsity regularizer defined in~(\ref{groupeq}), $\|.\|_{\text{F}}$ is the Frobenius norm, $\Yb^c=[\yb_1^c,\ldots,\yb_N^c]$, and $\lambda_i$ depends on the group size.
As explained in Section~\ref{sec:related}, the role of the Group Lasso penalty is to encourage the codes $\alphab_j$ belonging to the same cluster to share the same sparsity pattern, see Figure~\ref{fig:sparse}.
For homogeneity reasons, we also consider the normalization factor $\lambda_i={\lambda}/{\sqrt{|S_i|}}$, as in~\cite{mairal2009non}.
Minimizing~(\ref{optim_pblm})  is easy with the ISTA method since we know how to compute the proximal operator of $\Psi$, which is described below:
\begin{lemma}[Proximal operator for the Group Lasso]
   Consider a matrix~$\Ub$ and call $\Zb=\prox_{\lambda\|.\|_{1,2}}[\Ub]$. Then, for all row $\Zb^j$ of $\Zb$,
   \begin{equation}
      \Zb^j = \max\left (1- \frac{ \lambda }{\|\Ub^j\|_2} ,0\right) \Ub^j.
   \end{equation}
\end{lemma}
Unfortunately, the procedure used to design the groups~$S_i$ does not yield a
differentiable relation between the denoised image~$\hat{\xb}$ and the
parameters to learn.  Therefore, we
relax the hard clustering assumption into a soft one, which is able to exploit
a similarity matrix~$\Sigmab$ representing pairwise relations between patches. Details about $\Sigmab$ are given in Section~\ref{sec:similarities}.
Yet, such a relaxation does not provide distinct groups of patches, preventing us from
using the Group Lasso penalty~(\ref{optim_pblm}).

This difficulty may be solved by introducing a joint relaxation of the Group Lasso penalty
and its proximal operator. First, we consider a similarity matrix~$\Sigmab$ that encodes the hard clustering assignment
used by LSSC---that is, $\Sigmab_{ij} = 1$ if $j$ is in~$S_i$ and~$0$ otherwise. Second, we note
that $\|\Ab_i\|_{1,2}=\|\Ab\diag(\Sigmab_i)\|_{1,2}$ where $\Sigmab_i$ is the $i$-th column of $\Sigmab$ that encodes
the $i$-th cluster membership.
Then, we adapt LISTA to problem~(\ref{optim_pblm}), with a different shrinkage parameter $\Lambdab^{(k)}_j$ per coordinate~$j$ and per iteration~$k$ as in Section~\ref{subsec:sc}, which yields
\begin{equation}
   \begin{split}
      \Bb & \leftarrow \Ab^{(k)} + \Cb^\top(\Yb^c-\Db\Ab^{(k)}) \\
      \Ab^{(k+1)}_{ij} & \leftarrow  \max\left (1- \frac{ \Lambda_j^{(k)} \sqrt{\|\Sigmab_i\|_1}}{\|(\Bb\diag(\Sigmab_i)^{\frac{1}{2}})^j\|_2} ,0\right) \Bb_{ij}, \\
   \end{split}\label{eq:lista_group}
\end{equation}
where the second update is performed for all $i,j$, the superscript $^j$ denotes the $j$-th row of a matrix, as above, and $\Ab_{ij}$ is simply the $j$-th entry of $\alphab_i$.

We are now in shape to relax the hard clustering assumption by allowing any
similarity matrix $\Sigmab$ in~(\ref{eq:lista_group}), leading to a relaxation of the Group Lasso penalty in Algorithm~\ref{alg:pseudo}. The resulting model is able to encourage similar patches to share similar sparsity patterns, while being trainable by minimization of the cost~(\ref{eq:cost}).

\subsection{Similarity Metrics}\label{sec:similarities}
We have computed similarities~$\Sigmab$ in various manners,
and implemented the following practical heuristics, which improve the computional complexity.

\paragraph{Online averaging of similarity matrices.}

As shown in Algorithm~\ref{alg:pseudo}, we use a convex combination of similarity matrices (using~$\nu_k$ in $[0,1]$, also learned by backpropagation),
which provides better results than computing the similarity on the current estimate only. This is expected since the current estimate~$\hat{\xb}$ may have lost too much signal information to compute accurately similarities,
whereas online averaging allows retaining information from the original signal.
We run an ablation study of our model reported in appendix to illustrate the need of similarity refinements during the iterations.
When they are no updates the model perfoms on average 0.15 dB lower than with 4 updates.

\vsp
\paragraph{Semi-local grouping.} As in all methods that exploit non-local self
similarities in images, we restrict the search for similar patches to $\yb_i$
to a window of size $w \times w$ centered around the patch. This approach is
commonly used to reduce the size of the similarity matrix and the global memory
cost of the method. This means that we will always have $\Sigmab_{ij}=0$ if pixels~$i$ and~$j$
are too far apart.

\vsp
\paragraph{Learned distance.}
We always use a similarity function of the
form~$\Sigmab_{ij}=e^{-{d_{ij}}}$, where $d_{ij}$ is a distance
between patches~$i$ and~$j$. As in classical deep learning models using non-local
approaches~\cite{liu2018non}, we do not directly use the $\ell_2$ distance
between patches.  Specifically, we consider
\begin{equation}
   d_{ij} = \|\diag(\kappab)(\hat{\xb}_i - \hat{\xb}_j)\|^2,
\end{equation}
where $\hat{\xb}_i$ and $\hat{\xb}_j$ are the $i$ and $j$-th patches from the current denoised image,
and $\kappab$ in $\Real^m$ is a set of weights, which are learned by backpropagation.

\subsection{Extension to Blind Denoising and Parameter Sharing}

The regularization parameter  $\lambda$  of Eq. (\ref{eq:l1_problem}) depends on the noise level.
In a blind denoising setting, it is possible to learn a shared set of dictionnaries $\{\Db,\Cb,\Wb\}$
and a set of different regularization parameters $\{\Lambda_{\sigma_0}, \dots, \Lambda_{\sigma_n} \}$ for various noise intensities.
At inference time, we use first a noise estimation algorithm from \cite{liu2013single} and then select the best regularization parameter to restore the image.

\subsection{Extension to Demosaicking}
Most modern digital cameras acquire color images by measuring only one color
channel per pixel, red, green, or blue, according to a specific pattern called
the Bayer pattern.  Demosaicking is the processing step that reconstruct a full
color image given these incomplete measurements. 

Originally addressed by using interpolation techniques~\cite{gunturk},
demosaicking has been successfully tackled by
sparse coding~\cite{mairal2009non} and deep learning models.
Most of them such as \cite{zhang2017learning,zhang2019rnan} rely on generic
architectures and black box models that do not encode a priori knowledge about the 
problem, whereas the authors of \cite{kokkinos2019iterative} 
propose an iterative algorithm that relies on the physics of the acquisition process.
Extending our model to demosaicking (and in fact to other inpainting tasks with small holes)
can be achieved by introducing a mask~$\Mb_i$ in the formulation for unobserved pixel values.
Formally we define $\Mb_i$ for patch $i$ as a vector in $\{0,1\}^{m}$, and $\Mb = [\Mb_0, \dots, \Mb_N]$ in $\{0,1\}^{n\times N}$ represents all masks. 
Then, the sparse coding formulation becomes
\begin{equation} \label{mosaic_optim}
\min_{\Ab} \frac{1}{2} \| \Mb \odot (\Yb^c-\Db \Ab)\|_{\text{F}}^2 + \sum_{i=1}^N \Psi_i(\Ab),
\end{equation}
where $\odot$ denotes the elementwise product between two matrices.  The first updating rule of equation~(\ref{eq:lista_group}) is modified accordingly. This lead to a different update which has the effect of discarding reconstruction error of masked pixels,
\begin{equation}
   \Bb  \leftarrow \Ab^{(k)} +  \Cb^\top (\Mb \odot (\Yb^c-\Db\Ab^{(k)})). 
\end{equation}

\subsection{Practical variants and implementation}\label{sec:extensions}
Finally, we discuss other practical variants
and implementation details.

\vsp
\paragraph{Dictionary initialization.}
A benefit of designing an architecture with a sparse coding
interpretation, is that the parameters $\Db,\Cb,\Wb$ can be initialized with a
classical dictionary learning approach, instead of using random weights, which
makes the initialization robust.  To do so, we use SPAMS toolbox
\cite{mairal2010online}.

\vsp
\paragraph{Block processing and dealing with border effects.}
The size of the tensor $\Sigmab$ grows quadratically with the image size, which
requires processing sequentially image blocks. 
Here, the block size is chosen to match the size~$w$ of the non local
window, which requires taking into account two important details:

(i) Pixels close to the image border belong to fewer patches than those
from the center, and thus receive less estimates in the averaging procedure.
When processing images per block, it is thus important to have a small overlap
between blocks, such that the number of estimates per pixel is consistent across the image.

(ii) We also process image blocks for training. It then is important to
take border effects into account, by rescaling the loss by the number of pixel estimates.

\section{Experiments}\label{sec:exp}

\def \1{23}
\def \2{14}
\def \3{_07}
\def \4{20}

\def \width{2.0}
\def \Width{1.9}

\begin{figure*}[h]
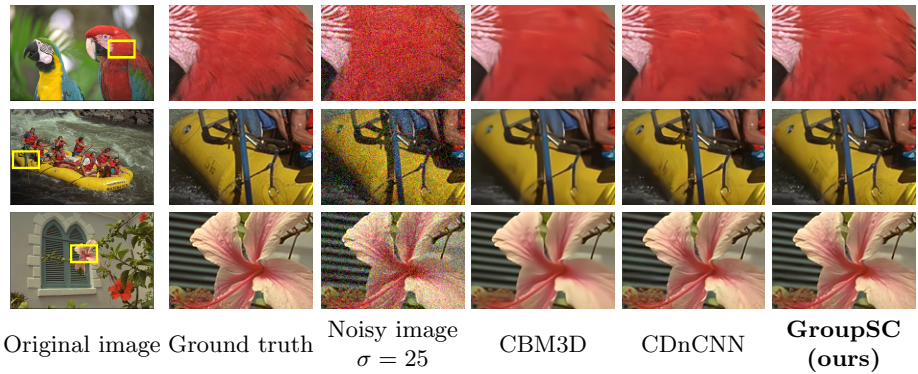

\centering
\begin{tabular}{ccccccc}
	\includegraphics[width=\Width cm]{figures/color_denoising/\1_bb.png}&
	\includegraphics[width=\Width cm]{figures/color_denoising/\1_gt.png}&
	\includegraphics[width=\Width cm]{figures/color_denoising/\1_noise.png}&
	\includegraphics[width=\Width cm]{figures/color_denoising/\1_bm3d.png}&
			\includegraphics[width=\Width cm]{figures/color_denoising/\1_dn.png}&
	\includegraphics[width=\Width cm]{figures/color_denoising/\1_group.png}
	\\
	
	\includegraphics[width=\Width cm]{figures/color_denoising/\2_bb.png}&
	\includegraphics[width=\Width cm]{figures/color_denoising/\2_gt.png}&
	\includegraphics[width=\Width cm]{figures/color_denoising/\2_noise.png}&
	\includegraphics[width=\Width cm]{figures/color_denoising/\2_bm3d.png}&
			\includegraphics[width=\Width cm]{figures/color_denoising/\2_dn.png}&
	\includegraphics[width=\Width cm]{figures/color_denoising/\2_group.png}
	\\	
	
	\includegraphics[width=\Width cm]{figures/color_denoising/\3_bb.png}&
	\includegraphics[width=\Width cm]{figures/color_denoising/\3_gt.png}&
	\includegraphics[width=\Width cm]{figures/color_denoising/\3_noise.png}&
	\includegraphics[width=\Width cm]{figures/color_denoising/\3_bm3d.png}&
		\includegraphics[width=\Width cm]{figures/color_denoising/\3_dn.png}&
	\includegraphics[width=\Width cm]{figures/color_denoising/\3_group.png}
	\\

Original image &	
	 Ground truth &
	\begin{tabular}{cc}Noisy image\\ $\sigma=25$  \end{tabular}&
	CBM3D & 
	CDnCNN &
	\begin{tabular}{cc}\textbf{GroupSC}\\ \textbf{(ours)} \end{tabular} \\
\end{tabular}

\caption{Color denoising results for 3 images from the Kodak24 dataset. Best seen in color
by zooming on a computer screen. More qualitative results for other tasks are in appendix.}	\label{fig:color}

\end{figure*}

\begin{table}[tb]
\footnotesize	
    \caption{\textbf{Blind denoising} on CBSD68, training on CBSD400. Performance is measured in terms of average PSNR. 
    SSIMs are in the appendix. 
    Best is in bold, second is underlined.}\label{tab:blind}
    \centering
    \begin{tabular}{ccccccc}
    \toprule
    \multirow{2}{*}{\begin{tabular}[c]{@{}l@{}}Noise\\level\end{tabular}} & CBM3D\cite{dabov2009bm3d} & CDnCNN-B \cite{zhang2017beyond} & CUNet\cite{lefkimmiatis2018universal} & CUNLnet\cite{lefkimmiatis2018universal} & SC (ours) & GroupSC (ours)  \\ 
     &-  & 666k & 93k & 93k & 115k  & 115k\\ \midrule
    5 & 40.24 & 40.11 & 40.31 & \underline{40.39} & 40.30 &  \textbf{40.43}\\
    10  & 35.88 & 36.11& 36.08&\underline{36.20} & 36.07 &  \textbf{36.29}\\
    15 & 33.49 & 33.88 & 33.78 & \underline{33.90}& {33.72} & \textbf{34.01}  \\
     20 &31.88 & \underline{32.36} & 32.21& {32.34}& 32.11& \textbf{32.41}\\
     25& 30.68 & \underline{31.22} & 31.03 & 31.17 & 30.91 & \textbf{31.25} \\    \bottomrule
    \end{tabular}

    \label{tab:my_label}
\end{table}

\begin{table}[tb]
\footnotesize
	\centering
   \caption{ \textbf{Color denoising} on CBSD68, training on CBSD400 for all methods except CSCnet (Waterloo+CBSD400). 
   Performance is measured in terms of average PSNR. SSIMs are reported in the appendix.}\label{colour_table}
\begin{tabular}{@{}lccccccccc@{}}
\toprule
\multirow{2}{*}{Method} &\multirow{2}{*}{Trainable } & \multirow{2}{*}{Params} & \multicolumn{4}{c}{Noise level ($\sigma$)} \\
& & & 5 & 10 & 15 & 25 & 30 & 50  \\
\midrule
CBM3D \cite{dabov2007image} & \xmark &-& 40.24 & - & 33.49& 30.68&-&27.36\\
\midrule
   CSCnet \cite{simon2019rethinking} &  & 186k & - &- &33.83 &31.18 &-&28.00 \\
CNLNet\cite{lefkimmiatis2017non}&  &-& -&-&33.69 & 30.96& - &27.64 \\
FFDNET \cite{zhang2018ffdnet}&  &486k & -&-&33.87 &31.21& - &27.96 \\
CDnCNN  \cite{zhang2017beyond}&  & 668k & 40.50 & 36.31 & 33.99 &31.31& -& 28.01\\
RNAN \cite{zhang2019rnan}&  & 8.96M &-& \textbf{36.60}&- & -&\textbf{30.73}&\textbf{28.35} \\
\midrule

{SC} (baseline) & & {119k} & {40.44}& - & {33.75} & 30.94 & - &27.39  \\
{GroupSC} (ours)& &{119k}&\underline{40.58}& \underline{36.40} &\underline{34.11} &\underline{31.44} & \underline{30.58} &\underline{28.05}  \\
\bottomrule
\end{tabular}
\end{table}

\begin{table}[tb]
\small
\centering
\caption{\textbf{Grayscale Denoising} on BSD68, training on BSD400 for all methods except CSCnet (Waterloo+BSD400).  
Performance is measured in terms of average PSNR.
SSIMs are reported in the appendix.}
\label{gray_table}
\begin{tabular}{@{}lcccccc@{}}
\toprule
\multirow{2}{*}{Method}  & \multirow{2}{*}{Trainable} & \multirow{2}{*}{Params}   & \multicolumn{4}{c}{Noise Level ($\sigma$)} \\
& & & 5 & 15 & 25 & 50 \\
\midrule
BM3D \cite{dabov2007image}& \xmark &-&37.57 &31.07& 28.57&25.62\\
LSSC \cite{mairal2009non}& \xmark &-&37.70&31.28&28.71 &25.72  \\
BM3D PCA \cite{dabov2009bm3d}& \xmark  & -&37.77 & 31.38  &28.82& 25.80\\
 \midrule
 TNRD \cite{chen2016trainable}&  &- &- & 31.42 & 28.92 & 25.97 \\
CSCnet \cite{simon2019rethinking} &  &62k & 37.84 & 31.57 & { 29.11} & 26.24  \\
   CSCnet(BSD400) \cite{simon2019rethinking}\footnotemark[2]  &  &62k & 37.69 & 31.40 & { 28.93} & 26.04  \\
   LKSVD~\cite{scetbon2019deep} &  & 45K & - & 31.54 & 29.07 & 26.13 \\
NLNet \cite{lefkimmiatis2017non}& & -&- & 31.52 & 29.03 &26.07 \\
FFDNet \cite{zhang2018ffdnet}&  & 486k&- & 31.63 & 29.19 &26.29 \\
DnCNN   \cite{zhang2017beyond}&  &556k & {37.68} &  \underline{31.73} & 29.22 & 26.23 \\
N3   \cite{plotz2018neural}&  & 706k & - &  - &  \underline{29.30} &  \underline{26.39}  \\
NLRN  \cite{liu2018non} &  & 330k& \underline{37.92} &\textbf{ 31.88} &\textbf{ 29.41} &\textbf{ 26.47}  \\
\midrule
{SC} (baseline)& &  {68k}& 37.84 & 31.46&28.90&25.84 \\ 
{GroupSC} (ours)&   & {68k}&\textbf{37.95}&31.71& 29.20& 26.17 \\
\bottomrule
\end{tabular}
\end{table}

\begin{table}

    \caption{ \textbf{Jpeg artefact reduction} on Classic5 with training on CBSD400. 
    Performance is measured in terms of average PSNR. 
    SSIMs are reported in the appendix.}
\footnotesize
    \centering
    \begin{tabular}{cccccccc}
    \toprule
    \begin{tabular}[c]{@{}c@{}}Quality\\ factor\end{tabular} &jpeg & SA-DCT \cite{foi2007pointwise} & AR-CNN \cite{yu2016deep} & TNRD\cite{chen2016trainable} & DnCNN-3 \cite{zhang2017beyond} & SC & GroupSC  \\
    \midrule
    qf = 10 &  27.82 & 28.88 &29.04 & 29.28 &\underline{29.40} & 29.39& \textbf{29.61}\\
    qf = 20 &  30.12 & 30.92 &31.16 & 30.12& \underline{31.63} & {31.58} &\textbf{31.78}\\
    qf = 30 & 31.48  & 32.14 &32.52 & 31.47 & \underline{32.91}  &  {32.80} &\textbf{33.06}  \\
    qf = 40 & 32.43  & 33.00 &33.34 & - &\underline{33.75} & 33.75&\textbf{33.91}\\
    \bottomrule
    \end{tabular}
    \label{tab:jpeg}
\end{table}
\footnotetext{We run here the model with the code provided by the authors online on the smaller training set BSD400.}

\begin{table} \label{mosaic_table}
\centering

    \footnotesize	
    \caption{\textbf{Demosaicking.} Training on CBSD400 unless a larger dataset is specified between parenthesis.
     Performance is measured in terms of average PSNR. SSIMs are reported in the appendix.}
    \begin{tabular}{@{}lccccc@{}}
    \toprule
    Method &Trainable & Params & Kodak24 &BSD68 & Urban100 \\
    \midrule
    LSSC     & \xmark   &  - & 41.39 & 40.44 & 36.63\\
    \midrule
    IRCNN \cite{zhang2017learning} (BSD400+Waterloo \cite{ma2016waterloo})&   & -  &  40.54  &39.9& 36.64 \\
    Kokinos \cite{kokkinos2018deep} (MIT dataset \cite{gharbi2016deep}) & & 380k &    41.5   & -& - \\
    MMNet \cite{kokkinos2019iterative} (MIT dataset \cite{gharbi2016deep})&  & 380k & 42.0 & - & -\\
    RNAN \cite{zhang2019rnan}&  & {8.96M} &  \textbf{42.86}  & \underline{42.61} & - \\
    \midrule
    {SC} (ours)&  & 119k& 42.34 & 41.88 & 37.50\\ 

    {GroupSC}   (ours)&  &  {119k}  &  \underline{42.71}& \textbf{42.91} & \textbf{38.21} \\ 
    \bottomrule
    \end{tabular}

\end{table}

\paragraph{Training details and datasets.} In our experiments, we adopt the
setting of \cite{zhang2017beyond}, which is the most standard one used by
recent deep learning methods, allowing a simple and fair comparison.  In
particular, we use as a training set a subset of the Berkeley Segmentation
Dataset (BSD) \cite{martin2001database}, called BSD400.
We evaluate our models on 3 popular benchmarks: BSD68 (with no overlap with BSD400), Kodak24, and Urban100  \cite{huang2015single} and on Classic5 for Jpeg deblocking, following \cite{foi2007pointwise,yu2016deep}.
For gray denoising and Jpeg deblocking  we choose a patch size of $9\times9$ and dictionary with 256 atoms for our models, whereas we choose a patch size of  $7\times7$ for color denoising and demosaicking.
 For all our experiments, we randomly extract patches of size $56\times56$ whose size equals the neighborhood for non-local operations and optimize the parameters of our models using ADAM~\cite{kingma2014adam}.  Similar to \cite{simon2019rethinking}, we normalize the initial dictionnary $\Db_0$ by its largest singular value, which helps the LISTA algorithm to converge. We also implemented a backtracking strategy that automatically decreases the learning rate by a factor 0.5 when the training loss diverges.
Additional training details can be found in the appendix for reproductibility purposes.

\vsp
\paragraph{Performance measure.} We use the PSNR as a quality measure, but SSIM scores for our experiments are provided in the appendix, leading to similar conclusions.

\vsp
\paragraph{Grayscale Denoising.}
We train our models under the same setting as \cite{zhang2017beyond,lefkimmiatis2017non,liu2018non}. We corrupt images with synthetic additive gaussian noise with a variance $\sigma = \{5,15,25,50 \}$ and train a different model for each~$\sigma$ and report the performance in terms of PSNR. 
Our method appears to perform on par with DnCNN for $\sigma \geq 10$ and performs significantly better for low-noise settings. Finaly we  provide results on other datasets in the appendix. 
On BSD68 the light version of our method runs 10 times faster than NLRN \cite{liu2018non} (2.17s for groupSC and 21.02s for NLRN), see the appendix for detailed experiments concerning the running time our our method ans its variants.

\vsp
\paragraph{Color Image Denoising}
We train our models under the same setting as \cite{lefkimmiatis2017non,zhang2017beyond}; we corrupt images with synthetic additive gaussian noise with a variance $\sigma = \{5,10,15,25,30,50\} $ and we train a different model for each variance of noise. 
%
For reporting both qualitative and quantitative results of {BM3D-PCA} \cite{dabov2009bm3d} and {DnCNN} \cite{zhang2017beyond} we used the implementation realeased by the authors. For the other methods we provide the numbers reported in the corresponding papers.
We report the performance of our model in Table \ref{colour_table} and report qualitative results in Figure~\ref{fig:color}, along with those of competitive approaches, and provide results on other datasets in the appendix. 
Overall, it seems that RNAN performs slightly better than GroupSC, at a cost of using $76$ times more parameters.

\vsp
\paragraph{Blind Color Image Denoising.}
We compare our model with \cite{lefkimmiatis2018universal,zhang2017beyond,dabov2009bm3d} and report our results in Table \ref{tab:blind}. \cite{lefkimmiatis2018universal} trains two different models in the range [0,25] and [25,50]. We compare with their model trained in the range [0,25] for a fair comparaison. We use the same hyperparameters than the one used for color denoising experiments. Our model performs consistently better than other methods.

\vsp
\paragraph{Demosaicking.} \label{section_expe_mosaic}
We follow the same experimental setting as IRCNN \cite{zhang2017learning}, but we do not crop the output images similarly to \cite{zhang2017learning,mairal2009non} since \cite{zhang2019rnan} does not seem to perform such an operation according to their code online. We compare our model with sate-of-the-art deep learning methods \cite{kokkinos2018deep,kokkinos2019iterative,zhang2019rnan} and also report the performance of LSSC.  
For the concurrent methods we provide the numbers reported in the corresponding papers. 
On  BSD68, the light version of our method(groupsc) runs at about the same speed than RNAN for demosaicking (2.39s for groupsc and 2.31s for RNAN).
We observe that our baseline provides already very good results, which is surprising given its simplicity, but suffers from more visual artefacts than GroupSC (see Fig. \ref{fig:mosaic}). Compared to RNAN, our model is much smaller
and shallower (120 layers for RNAN and 24 iterations for ours). We also note that CSR performs poorly in comparison with groupSC. 

\vsp
\paragraph{Compression artefacts reduction.}
For jpeg deblocking, we compare our approach with
state-of-the-art methods using the same experimental setting: we only
restore images in the Y channel (YCbCr space) and train our models on
the CBSD400 dataset. Our model performs consistently better than other approaches.

\section{Conclusion}
We have presented a differentiable algorithm based on non-local sparse image
models, which performs on par or better than recent deep learning models, while
using significantly less parameters.  We believe that the performance of such
approaches---including the simple SC baseline---is surprising given the small
model size, and given the fact that the algorithm can be interpreted as a
single sparse coding layer operating on fixed-size patches.
This observation paves the way for future work for sparse coding models that should 
be able to model the local stationarity of natural images at multiple scales,
which we expect should perform even better.
We believe that our work also confirms that model-based image restoration principles  
developed about a decade ago are still useful to improve current deep learning models 
and are a key to push their current limits.

\section*{Acknowledgements}

JM and BL were supported by the ERC grant number 714381 (SOLARIS project)
and by ANR 3IA MIAI@Grenoble Alpes (ANR-19-P3IA-0003). 
JP was supported in part by the Louis Vuitton/ENS chair in artificial intelligence and the Inria/NYU collaboration.
In addition, this work was funded in part by the French government under management of Agence Nationale de la Recherche
as part of the "Investissements d'avenir" program, reference ANR-19-P3IA-0001 (PRAIRIE 3IA Institute)
and was performed using HPC resources from GENCI–IDRIS (Grant 2020-AD011011252).

\bibliographystyle{splncs04}
\bibliography{reference}

\clearpage
\appendix
\section*{Appendix}

This supplementary material is organized as follows:
In section~\ref{sec:csr}, we provide a brief description of an other approach to take into account self similarities in sparse models.
In Section~\ref{sec:impl}, we provide implementation details that are useful to reproduce the results of our paper (note that the code is also provided).
In Section~\ref{sec:res1}, we present additional quantitative results that were not included in the main paper for space limitation reasons;
we notably provide the SSIM quality metric~\cite{wang2003multiscale} for grayscale, color, and demosaicking experiments; the SSIM score is sometimes more meaningful than PSNR (note
that the conclusions presented in the main paper remain unchanged, except for grey image denoising, where our method becomes either closer or better than NLRN, whereas
it was slightly behind in PSNR); we also present ablation studies and provide additional baselines for demosaicking and denoising.
Section~\ref{sec:proofs} is devoted to the proof of Proposition 1, and finally in Section~\ref{sec:res2}, we present additional qualitative results (which require zooming on a computer screen).
Finally, in section~\ref{sec:viz} we included Visualizations of parameters learned by our model to provide better intuition regarding our approach.

\section{Centralised Sparse Representation}\label{sec:csr}

\vsp
A different approach to take into account self similarities in sparse models is
the CSR approach of \cite{dong2012nonlocally}. This approach
is easier to turn into a differentiable algorithm than the LSSC method, but we have
empirically observed that it does not perform as well. Nevertheless, we believe it to be
conceptually interesting, and we provide a brief description below.
The idea consists of regularizing each code $\alphab_i$ with the function
\begin{equation}
    \label{eq:csreq}
    \Psi_i(\alphab_i) =  \|\alphab_i\|_1 + \gamma \|\alphab_i - \betab_i\|_1,
\end{equation}
where $\betab_i$ is obtained by a weighted average of prevous codes.
Specifically, given some codes~$\alphab_i^{(k)}$ obtained at iteration~$k$ and a similarity matrix~$\Sigmab$, we compute
\begin{equation}
    \betab_i^{(k)} = \sum_{j} \frac{\Sigmab_{ij}}{\sum_{l} \Sigmab_{il}}\alphab_j^{(k)},
\end{equation}
and the weights $\betab_i^{(k)}$ are used in~(\ref{eq:csreq}) in order to compute the codes $\alphab_i^{(k+1)}$.
Note that the original CSR method of~\cite{dong2012nonlocally} uses similarities of the form
$\Sigmab_{ij} =  \exp{\left(-\frac{1}{2\sigma^2} \| \Wb\alphab_i-\Wb\alphab_j\|_2^2 \right)} $, but other similarities functions may be used.

Even though~\cite{dong2012nonlocally} does not use a proximal gradient descent
method to solve the problem regularized with~(\ref{eq:csreq}), the next proposition shows
that it admits a closed form, which is a key to turn CSR into a differentiable algorithm.
To the best of our knowledge,
this expression is new; its proof is given in the~appendix.

\begin{proposition}[Proximal operator of the CSR penalty]\label{csr_prox}
    Consider $\Psi_i$ defined in~(\ref{eq:csreq}). Then, for all $\ub$ in~$\Real^p$,

    \begin{equation*}
        \prox_{\lambda \Psi_i}[\ub] = S_\lambda \big( S_{\lambda \gamma} \left (\ub  - \betab_i - \lambda \sign(\betab_i) \right ) \\ +\betab_i + \lambda\sign(\betab_i) \big ),
    \end{equation*}
    where $S_\lambda$ is the soft-thresholding operator, see Figure~\ref{fig:prox_csr}.
\end{proposition}

\noindent
\begin{minipage}{0.56\textwidth}
    Despite the apparent complexity of the formula, it remains a continuous function of the input and is differentiable almost everywhere, hence compatible with end-to-end training.
    Qualitatively, the shape of the proximal mapping has a simple interpretation. It pulls codes either to zero, or to the code weighted average $\betab_i$.

\end{minipage}
\hfill
\begin{minipage}{0.42\textwidth}

    \centering\raisebox{\dimexpr \topskip-\height}{%
        \includegraphics[width=\textwidth]{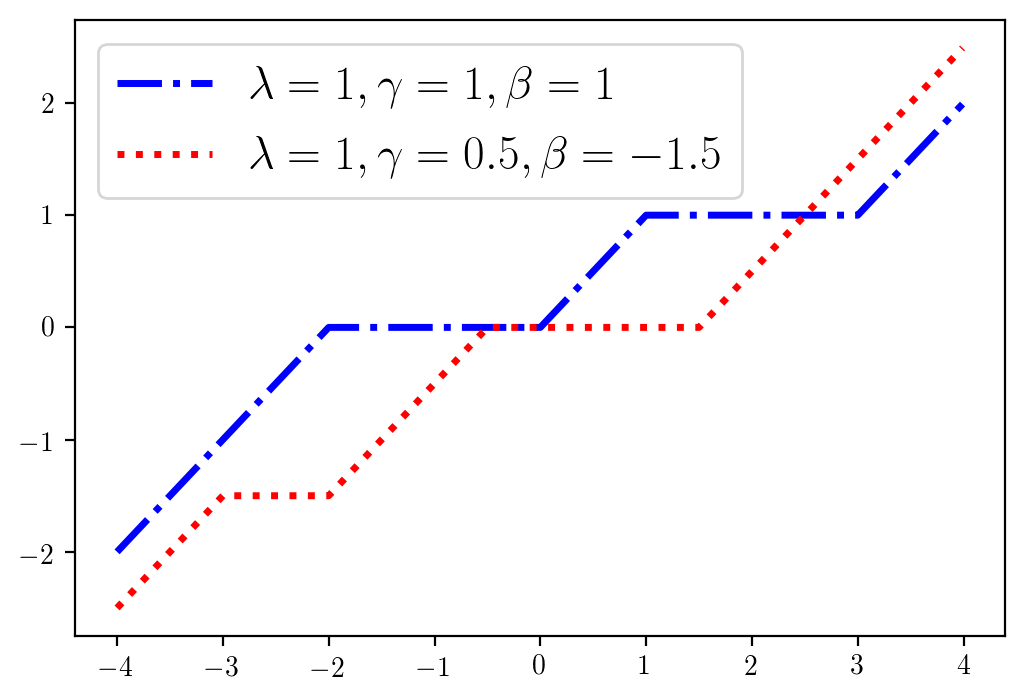}}
    \captionof{figure}{$ \prox_{\lambda \Psi_i}$ for various $\lambda, \gamma, \beta$}\label{fig:prox_csr}
\end{minipage}

At each iteration, the similarity matrix is updated along with the codes $\betab_i$. The proximal operator can then easily be plugged into our framework.
We reported performance of the CSR approach in the main paper for grayscale denoising, color denoising and demosaicking.
Performance of the CSR approach are reported in Tables \ref{colour_table_csr}, \ref{gray_table_csr}, \ref{mosaic_table_csr}.
We observe that it performs significantly
better than the baseline SC but is not as effective as GroupSC overall.

\begin{table}[htb]
    \footnotesize
    \centering
    \caption{ \textbf{Color denoising} on CBSD68, training on CBSD400 for all methods except CSCnet (Waterloo+CBSD400).
        Performance is measured in terms of average PSNR. SSIMs are reported in the appendix.}\label{colour_table_csr}
    \begin{tabular}{@{}lccccccccc@{}}
        \toprule
        \multirow{2}{*}{Method}           & \multirow{2}{*}{Trainable } & \multirow{2}{*}{Params} & \multicolumn{4}{c}{Noise level ($\sigma$)}                                                                                                     \\
                                          &                             &                         & 5                                          & 10                & 15                & 25                & 30                & 50                \\
        \midrule
        CBM3D \cite{dabov2007image}       & \xmark                      & -                       & 40.24                                      & -                 & 33.49             & 30.68             & -                 & 27.36             \\
        \midrule
        CSCnet \cite{simon2019rethinking} &                             & 186k                    & -                                          & -                 & 33.83             & 31.18             & -                 & 28.00             \\
        CNLNet\cite{lefkimmiatis2017non}  &                             & -                       & -                                          & -                 & 33.69             & 30.96             & -                 & 27.64             \\
        FFDNET \cite{zhang2018ffdnet}     &                             & 486k                    & -                                          & -                 & 33.87             & 31.21             & -                 & 27.96             \\
        CDnCNN  \cite{zhang2017beyond}    &                             & 668k                    & 40.50                                      & 36.31             & 33.99             & 31.31             & -                 & 28.01             \\
        RNAN \cite{zhang2019rnan}         &                             & 8.96M                   & -                                          & \textbf{36.60}    & -                 & -                 & \textbf{30.73}    & \textbf{28.35}    \\
        \midrule
        {SC} (baseline)                   &                             & {119k}                  & {40.44}                                    & -                 & {33.75}           & 30.94             & -                 & 27.39             \\
        {CSR} (ours)                      &                             & {119k}                  & {40.53}                                    & -                 & {34.05}           & {31.33}           & -                 & {28.01}           \\
        {GroupSC} (ours)                  &                             & {119k}                  & \underline{40.58}                          & \underline{36.40} & \underline{34.11} & \underline{31.44} & \underline{30.58} & \underline{28.05} \\
        \bottomrule
    \end{tabular}
\end{table}

\begin{table}[htb]
    \small
    \centering
    \caption{\textbf{Grayscale Denoising} on BSD68, training on BSD400 for all methods except CSCnet (Waterloo+BSD400).
        Performance is measured in terms of average PSNR.
        SSIMs are reported in the appendix.}
    \label{gray_table_csr}
    \begin{tabular}{@{}lcccccc@{}}
        \toprule
        \multirow{2}{*}{Method}                                   & \multirow{2}{*}{Trainable} & \multirow{2}{*}{Params} & \multicolumn{4}{c}{Noise Level ($\sigma$)}                                                             \\
                                                                  &                            &                         & 5                                          & 15                & 25                & 50                \\
        \midrule
        BM3D \cite{dabov2007image}                                & \xmark                     & -                       & 37.57                                      & 31.07             & 28.57             & 25.62             \\
        LSSC \cite{mairal2009non}                                 & \xmark                     & -                       & 37.70                                      & 31.28             & 28.71             & 25.72             \\
        BM3D PCA \cite{dabov2009bm3d}                             & \xmark                     & -                       & 37.77                                      & 31.38             & 28.82             & 25.80             \\
        \midrule
        TNRD \cite{chen2016trainable}                             &                            & -                       & -                                          & 31.42             & 28.92             & 25.97             \\
        CSCnet \cite{simon2019rethinking}                         &                            & 62k                     & 37.84                                      & 31.57             & { 29.11}          & 26.24             \\
        CSCnet(BSD400) \cite{simon2019rethinking}\footnotemark[2] &                            & 62k                     & 37.69                                      & 31.40             & { 28.93}          & 26.04             \\
        LKSVD~\cite{scetbon2019deep}                              &                            & 45K                     & -                                          & 31.54             & 29.07             & 26.13             \\
        NLNet \cite{lefkimmiatis2017non}                          &                            & -                       & -                                          & 31.52             & 29.03             & 26.07             \\
        FFDNet \cite{zhang2018ffdnet}                             &                            & 486k                    & -                                          & 31.63             & 29.19             & 26.29             \\
        DnCNN   \cite{zhang2017beyond}                            &                            & 556k                    & {37.68}                                    & \underline{31.73} & 29.22             & 26.23             \\
        N3   \cite{plotz2018neural}                               &                            & 706k                    & -                                          & -                 & \underline{29.30} & \underline{26.39} \\
        NLRN  \cite{liu2018non}                                   &                            & 330k                    & \underline{37.92}                          & \textbf{ 31.88}   & \textbf{ 29.41}   & \textbf{ 26.47}   \\
        \midrule
        {SC} (baseline)                                           &                            & {68k}                   & 37.84                                      & 31.46             & 28.90             & 25.84             \\
        {CSR} (ours)                                              &                            & {68k}                   & 37.88                                      & 31.64             & 29.16             & 26.08             \\
        {GroupSC} (ours)                                          &                            & {68k}                   & \textbf{37.95}                             & 31.71             & 29.20             & 26.17             \\
        \bottomrule
    \end{tabular}
\end{table}

\begin{table}[htb]
    \centering
    \footnotesize
    \caption{\textbf{Demosaicking.} Training on CBSD400 unless a larger dataset is specified between parenthesis.
        Performance is measured in terms of average PSNR. SSIMs are reported in the appendix.} \label{mosaic_table_csr}
    \begin{tabular}{@{}lccccc@{}}
        \toprule
        Method                                                                 & Trainable & Params  & Kodak24           & BSD68             & Urban100       \\
        \midrule
        LSSC                                                                   & \xmark    & -       & 41.39             & 40.44             & 36.63          \\
        \midrule
        IRCNN \cite{zhang2017learning} (BSD400+Waterloo \cite{ma2016waterloo}) &           & -       & 40.54             & 39.9              & 36.64          \\
        Kokinos \cite{kokkinos2018deep} (MIT dataset \cite{gharbi2016deep})    &           & 380k    & 41.5              & -                 & -              \\
        MMNet \cite{kokkinos2019iterative} (MIT dataset \cite{gharbi2016deep}) &           & 380k    & 42.0              & -                 & -              \\
        RNAN \cite{zhang2019rnan}                                              &           & {8.96M} & \textbf{42.86}    & \underline{42.61} & -              \\
        \midrule
        {SC} (ours)                                                            &           & 119k    & 42.34             & 41.88             & 37.50          \\
        CSR (ours)                                                             &           & 119k    & 42.25             & -                 & -              \\
        {GroupSC}   (ours)                                                     &           & {119k}  & \underline{42.71} & \textbf{42.91}    & \textbf{38.21} \\
        \bottomrule
    \end{tabular}

\end{table}

\section{Implementation Details and Reproducibility}\label{sec:impl}

\paragraph{Training details.}
During training, we randomly extract patches $56\times56$ whose size equals the window size used for computing non-local self-similarities. We apply a mild data augmentation (random rotation by
$90^\circ$ and horizontal flips). We optimize the parameters of our models using ADAM~\cite{kingma2014adam} with a minibatch size of $32$.
All the models are trained for 300 epochs for denoising and demosaicking. The learning rate is set to $6 \times 10^{-4}$ at initialization and is sequentially lowered during training
by a factor of 0.35 every 80 training steps, in the same way for all experiments. Similar to \cite{simon2019rethinking}, we normalize the initial dictionary~$\Db_0$ by its largest singular value, which helps the LISTA
algorithm to converge faster. We initialize the matrices $\Cb$,$\Db$ and~$\Wb$ with the same value, similarly to the implementation of~\cite{simon2019rethinking} released by the authors.
\footnote{The implementation of CSCnet~\cite{simon2019rethinking} is available here \url{https://github.com/drorsimon/CSCNet/}.}
Since too large learning rates can make the model diverge (as for any neural network), we have implemented a backtracking strategy that automatically decreases the learning rate by a factor 0.8 when the loss function
increases too much on the training set, and restore a previous snapshot of the model. Divergence is monitored by computing the loss on the training set every 20 epochs.
Training the GroupSC model for color denoising takes about 2 days on a Titan RTX GPU.

\paragraph{Accelerating inference.}
In order to make the inference time of the non-local models faster, we do not update similarity maps at every step: we update patch similarities every $1/f$ steps, where $f$ is the frequency of the correlation updates.
We summarize in Table \ref{hyperparam} the set of hyperparameters that we selected for the experiments reported in the main tables.

\begin{table}[h!]
    \small
    \centering
    \caption{Hyper-parameters chosen for every task.}
    \label{hyperparam}
    \begin{tabular}{lcccc}
        \toprule
        Experiment                    & Color denoising & Gray denoising & Demosaicking & Jpeg Deblocking \\
        \midrule
        Patch size                    & 7               & 9              & 7            & 9               \\
        Dictionary size               & 256             & 256            & 256          & 256             \\
        Nr epochs                     & 300             & 300            & 300          & 300             \\
        Batch size                    & 32              & 32             & 32           & 32              \\
        $K$ iterations                & 24              & 24             & 24           & 24              \\
        Middle averaging              & \cmark          & \cmark         & \cmark       & \cmark          \\

        \begin{tabular}[c]{@{}l@{}}Correlation update\\ frequency $f$\end{tabular} & ${1}/{6}$       & ${1}/{6}$      & ${1}/{8}$    & ${1}/{6}$       \\
        \bottomrule
    \end{tabular}

\end{table}

\section{Additional Quantitative Results and Ablation Studies}\label{sec:res1}

\subsection{Results on Other Datasets and SSIM Scores}

We provide additional grayscale denoising results of our model on the datasets BSD68, Set12, and Urban100 in terms of PSNR and SSIM in Table \ref{gray_app}.
Then, we present additional results for color denoising in Table~\ref{color_app}, for demosaicking in Table~\ref{fig:supp_mosa}, and for jpeg artefact reduction in Table~\ref{tab:jpeg}.
Note that we report SSIM scores for baseline methods, either because they report SSIM in the corresponding papers, or by running
the code released by the authors.

\begin{table}[h!]
    \small
    \centering
    \caption{\textbf{Grayscale denoising} results on different datasets. Training is performed on BSD400.  Performance is measured in terms of average PSNR (left number) and SSIM (right number). }
    \label{gray_app}
    \begin{tabular}{lccccc}
        \toprule
        Dataset & Noise & BM3D         & \begin{tabular}{cc}DnCNN\\556k \end{tabular}       & \begin{tabular}{cc}NLRN\\330k\end{tabular} & \begin{tabular}{cc} GroupSC\\ 68k\end{tabular}       \\
        \midrule
        \multirow{3}{*}{\textbf{Set12}}
                & 15    & 32.37/0.8952 & \underline{32.86}/0.9031             & \textbf{ 33.16/0.9070}         & 32.85/\underline{0.9063 }            \\
                & 25    & 29.97/0.8504 & 30.44/0.8622                         & \textbf{30.80/0.8689}          & \underline{30.44}/\underline{0.8642} \\
                & 50    & 26.72/0.7676 & \underline{27.18}/\underline{0.7829} & \textbf{27.64/0.7980}          & 27.14/0.7797                         \\
        \midrule
        \multirow{3}{*}{\textbf{BSD68}}
                & 15    & 31.07/0.8717 & \underline{31.73}/0.8907             & \textbf{31.88}/0.8932          & 31.70/\textbf{0.8963}                \\
                & 25    & 28.57/0.8013 & \underline{29.23}/0.8278             & \textbf{29.41}/0.8331          & 29.20/\textbf{0.8336}                \\
                & 50    & 25.62/0.6864 & \underline{26.23}/0.7189             & \textbf{26.47/0.7298}          & 26.18/0.7183                         \\
        \midrule
        \multirow{3}{*}{\textbf{Urban100}}
                & 15    & 32.35/0.9220 & 32.68/0.9255                         & \textbf{33.45/0.9354}          & \underline{32.72}/0.9308             \\
                & 25    & 29.70/0.8777 & 29.91/0.8797                         & \textbf{30.94/0.9018}          & \underline{30.05}/0.8912             \\
                & 50    & 25.95/0.7791 & 26.28/0.7874                         & \textbf{ 27.49/0.8279}         & \underline{26.43}/\underline{0.8002} \\ \bottomrule
    \end{tabular}
\end{table}

\begin{table}[h!]
    \small
    \centering
    \caption{\textbf{Color denoising} results on different datasets. Training is performed on CBSD400.  Performance is measured in terms of average PSNR (left number) or SSIM (right number). }
    \label{color_app}
    \begin{tabular}{lccc}
        \toprule
        Dataset & Noise & \begin{tabular}{cc}CDnCNN\\668k \end{tabular}       & \begin{tabular}{cc}GroupSC\\119k \end{tabular} \\ 	\midrule
        \multirow{4}{*}{\textbf{Kodak24}}
                & 15    & \underline{34.84}/\underline{0.9233} & \textbf{35.00/0.9275 }         \\
                & 25    & \underline{32.34}/\underline{0.8812} & \textbf{32.51/0.8867 }         \\
                & 50    & \underline{29.15}/\underline{0.7985} & \textbf{29.19/0.7993 }         \\
        \midrule	\multirow{4}{*}{\textbf{CBSD68}}
                & 15    & \underline{33.98}/\underline{0.9303} & \textbf{34.11}/\textbf{0.9353} \\
                & 25    & \underline{31.31}/\underline{0.8848} & \textbf{31.44}/\textbf{0.8917} \\
                & 50    & \underline{28.01}/\underline{0.7925} & \textbf{28.05}/\textbf{0.7974} \\
        \midrule	\multirow{4}{*}{\textbf{Urban100}}
                & 15    & \underline{34.11}/\underline{0.9436} & \textbf{34.14}/\textbf{0.9461} \\
                & 25    & \underline{31.66}/\underline{0.9145} & \textbf{31.69}/\textbf{0.9178} \\
                & 50    & \underline{28.16}/\underline{0.8410} & \textbf{28.23}/\textbf{0.8513} \\ 	\bottomrule
    \end{tabular}
\end{table}

\begin{table}[h!]

    \caption{ \textbf{Jpeg artefact reduction} on Classic5 with training on CBSD400. Performance is measured in terms of average PSNR.}
    \footnotesize
    \centering
    \begin{tabular}{ccccc}
        \toprule
        \begin{tabular}[c]{@{}c@{}}Quality\\ factor\end{tabular} & AR-CNN \cite{yu2016deep} & TNRD\cite{chen2016trainable} & DnCNN-3 \cite{zhang2017beyond}       & GroupSC                         \\
        \midrule
        10                             & 29.04/0.7929             & 29.28/0.7992                 & \underline{29.40}/\underline{0.8026} & \textbf{29.61}/ \textbf{0.8166} \\
        20                             & 31.16/0.8517             & 31.47/0.8576                 & \underline{31.63}/\underline{0.8610} & \textbf{31.78}/ \textbf{0.8718} \\
        30                             & 32.52/0.8806             & 32.78/0.8837                 & \underline{32.91}/\underline{0.8861} & \textbf{33.06}/ \textbf{0.8959} \\
        40                             & 33.34/0.8953             & -                            & \underline{33.75}/0.9003             & \textbf{33.91}/ \textbf{0.9093} \\
        \bottomrule
    \end{tabular}
    \label{tab:jpeg}
\end{table}

\begin{table}[h!] \label{supp_mosa}
    \centering

    \footnotesize
    \caption{\textbf{Demosaicking} results. Training on CBSD400 unless a larger dataset is specified between parenthesis. Performance is measured in terms of average PSNR (left) and SSIM (right).}\label{mosa_ssim}
    \begin{tabular}{@{}lccccc@{}}
        \toprule
        Method                      & Params & \textbf{Kodak24}                     & \textbf{BSD68}                       & \textbf{Urban100}                    \\
        \midrule

        IRCNN (BSD400+Waterloo)     & 107k   & \underline{40.54}/\underline{0.9807} & \underline{39.96}/\underline{0.9850} & \underline{36.64}/\underline{0.9743} \\
        {GroupSC} (CBSD400)  (ours) & {118k} & \textbf{42.71/0.9901}                & \textbf{42.91/0.9938}                & \textbf{38.21/0.9804}                \\

        \bottomrule
    \end{tabular}
    \vspace*{-0.5cm}

\end{table}

\subsection{Inference Speed and Importance of Similarity Refinements}

In table \ref{tab:speed},
we provide  a comparison of our model in terms of speed. We compare our model for demosaicking and color denoising with the methods NLRN.
This study shows how to balance the trade-off between speed and accuracy. Whereas the best model in accuracy achieves 31.71dB in PSNR with about 30s per image, a ``light'' version can
achieve 31.67dB in only 2.35s per image.
This ablation study also illustrates the need of similarity refinements during the iterations.
When they are no updates the model perfoms on average 0.15 dB lower than with 4 updates.

\begin{table}[h!]
    \caption{\textbf{Inference time (s)} per image / PSNR (in dB) for gray denoising task with $\sigma=15$, computed on BSD68. Inference time is measured using a Titan RTX gpu.}\label{tab:speed}
    \footnotesize
    \centering
    \begin{tabular}{@{}|c|c||c|c|c|c|@{}}
        \hline
        \multirow{2}{*}{\begin{tabular}[c]{@{}c@{}}Middle \\ averaging (6) \end{tabular}} & \multirow{2}{*}{$f_{\hat{\Sigmab}}$} & \multicolumn{4}{c|}{Stride between image blocks}                                               \\ \cline{3-6}
                                                        &                                      & $s=56$                                           & $s=48$       & $s=24$       & $s=12$        \\ \hline \hline

        \multirow{4}{*}{\xmark
        }                                               & $\infty$                             & 1.30 / 31.29                                     & 1.75 / 31.57 & 6.00 / 31.58 & 22.57 / 31.59 \\
                                                        & 12                                   & 1.41 / 31.36                                     & 1.85 / 31.64 & 6.57 / 31.66 & 24.44 / 31.66 \\
                                                        & 8                                    & 1.51 / 31.37                                     & 2.90 / 31.65 & 7.06 / 31.68 & 26.05 / 31.68 \\
                                                        & 6                                    & 1.59 / 31.38                                     & 2.15 / 31.65 & 7.48 / 31.68 & 27.60 / 31.69 \\ \hline

        \multirow{4}{*}{\cmark
        }                                               & $\infty   $                          & 1.30 / 31.29                                     & 1.75 / 31.57 & 6.00 / 31.58 & 22.57 / 31.59 \\
                                                        & 12                                   & 1.45 / 31.36                                     & 1.95 / 31.65 & 6.82 / 31.66 & 25.40 / 31.67 \\
                                                        & 8                                    & 1.63 / 31.38                                     & 2.17 / 31.66 & 7.61 / 31.68 & 27.92 / 31.70 \\
                                                        & 6                                    & 1.77 / 31.39                                     & 2.35 / 31.67 & 8.25 / 31.69 & 30.05 / 31.71 \\ \hline

        NLRN                                            & 330k                                 & \multicolumn{4}{c|}{23.02 / 31.88}                                                             \\  \hline
    \end{tabular}
    \vspace*{-0.2cm}
\end{table}

\subsection{Influence of Patch and Dictionary Sizes}

We measure in Table \ref{ablation_table} the influence of the patch size and the dictionary size for grayscale image denoising.
For this experiment, we run  a lighter version of the model groupSC in order to accelerate the training. The batch size was decreased from 25 to 16, the
frequency of the correlation updates was decreased from $1/6$ to $1/8$ and the intermediate patches are not approximated with averaging.
These changes accelerate the training but lead to slightly lower performances when compared with the model trained in the standard setting.
As can be seen in the table, better performance can be obtained by using larger dictionaries, at the cost of more computation. Note that
all other experiments conducted in the paper use a dictionary size of 256. Here as well, a trade-off between speed/number of parameters and accuracy can be chosen
by changing this default value.

\begin{table}[h!]
    \small
    \centering
    \caption{\textbf{Influence of the dictionary size and the patch size }on the denoising performance.
        Grayscale denoising on BSD68. Models are trained on BSD400. Models are trained in a light setting to accelerate training.}
    \label{ablation_table}
    \begin{tabular}{|c|l|c|c|c|}
        \hline
        Noise ($\sigma$) & Patch size & n=128 & n=256 & 512   \\
        \hline
        \multirow{3}{*}{$5$}
                         & k=7        & 37.91 & 37.92 & -     \\
                         & k=9        & 37.90 & 37.92 & 37.96 \\
                         & k=11       & 37.89 & 37.89 & -     \\
        \hline
        \multirow{3}{*}{$15$}
                         & k=7        & 31.60 & 31.63 & -     \\
                         & k=9        & 31.62 & 31.67 & 31.71 \\
                         & k=11       & 31.63 & 31.67 & -     \\
        \hline
        \multirow{3}{*}{$25$}
                         & k=7        & 29.10 & 29.11 & -     \\
                         & k=9        & 29.12 & 29.17 & 29.20 \\
                         & k=11       & 29.13 & 29.18 & -     \\
        \hline
    \end{tabular}
\end{table}

\subsection{Number of Unrolled Iterations}
We also investigated the impact of the depth of the model on the performance.
To do so, we conducted a denoising experiment using the light version of our model with a model with various number of
unrolled steps.
When changing the depth from K=12, to 36, we only measure a difference of 0.02dB.

\begin{table}[h!]
    \centering
    \caption{\textbf{Influence  of the number of unrolled iterations}.Grayscale denoising on BSD68.
        Models are trained on BSD400. Models are trained in a light setting to accelerate training.}

    \begin{tabular}{|l|c|c|c|}
        \hline
        Model           & \multicolumn{3}{l|}{Unrolled iterations }                 \\ \hline
        SC              & 28.90                                     & 28.91 & 28.90 \\
        GroupSC (light) & 29.10                                     & 29.12 & 29.12 \\ \hline
    \end{tabular}
\end{table}

\section{Proof of Proposition 1}\label{sec:proofs}

The proximal operator of the function $\Psi_i(\ub) = \| \ub \|_1 + \gamma \| \ub - \betab_i \|_1$  for $\ub$ in~$\Real^p$ is defined as
\begin{equation*} \label{prox_psi}
    \prox_{\lambda \Psi_i} [\zb]=  \argmin_{\ub \in \Real^p} \frac{1}{2}\|\zb-\ub\|^2 + \lambda \|\ub\|_1 + \lambda \gamma \| \ub - \betab_i\|_1
\end{equation*}
The optimality condition for the previous problem is
$$ 0 \in \triangledown (\frac{1}{2} || \zb - \ub ||_2^2) + \partial (\lambda ||\ub||_1) + \partial (\lambda \gamma ||\ub - \betab_i||_1) $$
$$\Leftrightarrow 0 \in \ub - \zb + \lambda \partial ||\ub||_1 + \lambda \gamma  \partial ||\ub - \betab_i||_1  $$
We consider each component separately. We suppose that $\betab_i[j] \neq 0$, otherwise  $\Psi_i(\ub)[j]$ boils down to the $\ell_1$ norm. And we also suppose $\lambda,\gamma >0$.

Let us examine the first case where $u[j] = 0$. The subdifferential of the $\ell_1$ norm is the interval $[-1,1]$ and the optimality condition is
\begin{align*}
    0 \in \ub[j] - \zb[j] + [-\lambda,\lambda] + \lambda \gamma \sign(\ub[j]-\betab_i[j]) \\
    \Leftrightarrow \zb[j] \in [-\lambda,\lambda] -  \lambda \gamma \sign(\betab_i[j])
\end{align*}
Similarly if $\ub[j] = \betab_i[j]$
\begin{equation*}
    \zb[j] \in \betab_i[j] + \lambda \sign(\betab_i[j]) + [-\lambda \gamma,\lambda \gamma]
\end{equation*}

Finally let us examine the case where $u[j] \neq 0$ and $u[j] \neq \betab_i[j]$: then, $\partial ||\ub||_1 = \sign(\ub[j])$ and $\partial ||\ub - \betab_i||_1 = \sign(\ub[j] - \betab_i[j])$. The minimum $u[j]^*$ is obtained as
\begin{align*}
    0 = \ub[j] - \zb[j] + \lambda \sign(\ub[j]) + \lambda \gamma \sign(\ub[j]-\betab_i[j]) \\
    \Leftrightarrow \ub[j]^* =  \zb[j] - \lambda \sign(\ub[j]^*) - \lambda \gamma \sign(\ub[j]^*-\betab_i[j])
\end{align*}

We study separately the cases where $\ub[j]>\betab[j]$, $0<\ub[j]<\betab[j]$ and $\ub[j]<0$ when $\betab_i[j]>0$ and proceed similarly when $\betab_i<0$. With elementary operations we can derive the expression of $\zb[j]$ for each case. Putting the cases all together we obtain the formula.

\section{Additional Qualitative Results}\label{sec:res2}

We show qualitative results for jpeg artefact reduction, color denoising, grayscale denoising, and demosaicking in Figures
\ref{fig:supp_color},
\ref{fig:supp_gray},
\ref{fig:supp_mosa}, respectively.

\def \1{barbara}
\def \2{lena}
\def \3{baboon}

\def \width{1.8}
\def \Width{1.75}

\begin{figure*}[h!]
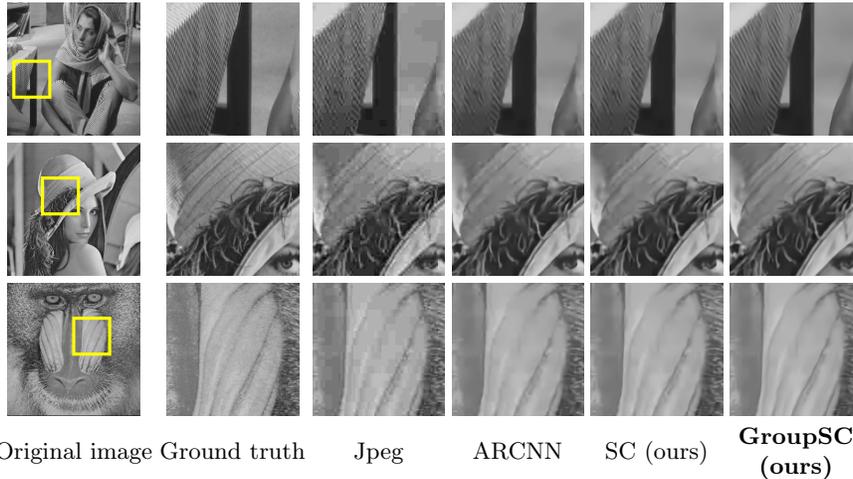

\centering
\begin{tabular}{cccccc}
	\includegraphics[width=\Width cm]{figures/jpeg/\1_bb.png}&
	\includegraphics[width=\Width cm]{figures/jpeg/\1_gt.png}&
    \includegraphics[width=\Width cm]{figures/jpeg/\1_noisy.png}&
    \includegraphics[width=\Width cm]{figures/jpeg/\1_arcnn.png}&
	\includegraphics[width=\Width cm]{figures/jpeg/\1_sc.png}&
	\includegraphics[width=\Width cm]{figures/jpeg/\1_group.png}
	\\
	
	\includegraphics[width=\Width cm]{figures/jpeg/\2_bb.png}&
	\includegraphics[width=\Width cm]{figures/jpeg/\2_gt.png}&
    \includegraphics[width=\Width cm]{figures/jpeg/\2_noisy.png}&
    \includegraphics[width=\Width cm]{figures/jpeg/\2_arcnn.png}&
	\includegraphics[width=\Width cm]{figures/jpeg/\2_sc.png}&
	\includegraphics[width=\Width cm]{figures/jpeg/\2_group.png}
	\\	
	\includegraphics[width=\Width cm]{figures/jpeg/\3_bb.png}&
	\includegraphics[width=\Width cm]{figures/jpeg/\3_gt.png}&
    \includegraphics[width=\Width cm]{figures/jpeg/\3_noisy.png}&
    \includegraphics[width=\Width cm]{figures/jpeg/\3_arcnn.png}&
	\includegraphics[width=\Width cm]{figures/jpeg/\3_sc.png}&
	\includegraphics[width=\Width cm]{figures/jpeg/\3_group.png}
	\\
Original image &	
	 Ground truth &
    Jpeg&
    ARCNN&
    SC (ours) &
	\begin{tabular}{cc}\textbf{GroupSC}\\ \textbf{(ours)} \end{tabular} \\
\end{tabular}
\vspace*{-0.4cm}
\caption{Jpeg artefact reduction results for 2 images from the Classic5 dataset. Best seen in color
by zooming on a computer screen.}	\label{fig:supp_jpeg}

\end{figure*}

\def \1{12}
\def \2{15}
\def \3{06}
\def \4{07}
\def \width{2.0}
\def \Width{1.9}

\begin{figure*}[h!]
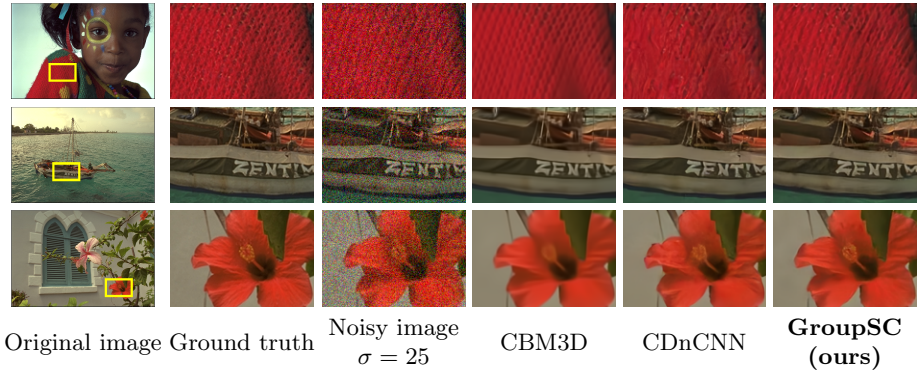

\centering
\begin{tabular}{ccccccc}
	
	\includegraphics[width=\Width cm]{figures/color_denoising/\2_bb.png}&
	\includegraphics[width=\Width cm]{figures/color_denoising/\2_gt.png}&
	\includegraphics[width=\Width cm]{figures/color_denoising/\2_noise.png}&
	\includegraphics[width=\Width cm]{figures/color_denoising/\2_bm3d.png}&
			\includegraphics[width=\Width cm]{figures/color_denoising/\2_dn.png}&
	\includegraphics[width=\Width cm]{figures/color_denoising/\2_group.png}
	\\	
	
	\includegraphics[width=\Width cm]{figures/color_denoising/\3_bb.png}&
	\includegraphics[width=\Width cm]{figures/color_denoising/\3_gt.png}&
	\includegraphics[width=\Width cm]{figures/color_denoising/\3_noise.png}&
	\includegraphics[width=\Width cm]{figures/color_denoising/\3_bm3d.png}&
		\includegraphics[width=\Width cm]{figures/color_denoising/\3_dn.png}&
	\includegraphics[width=\Width cm]{figures/color_denoising/\3_group.png}
	\\

	\includegraphics[width=\Width cm]{figures/color_denoising/\4_bb.png}&
	\includegraphics[width=\Width cm]{figures/color_denoising/\4_gt.png}&
	\includegraphics[width=\Width cm]{figures/color_denoising/\4_noise.png}&
	\includegraphics[width=\Width cm]{figures/color_denoising/\4_bm3d.png}&
		\includegraphics[width=\Width cm]{figures/color_denoising/\4_dn.png}&
	\includegraphics[width=\Width cm]{figures/color_denoising/\4_group.png}
	\\
%
Original image &	
	 Ground truth &
	\begin{tabular}{cc}Noisy image\\ $\sigma=25$  \end{tabular}&
	CBM3D & 
	CDnCNN &
	\begin{tabular}{cc}\textbf{GroupSC}\\ \textbf{(ours)} \end{tabular} \\
\end{tabular}

\caption{Color denoising results for 3 images from the Kodak24 dataset. Best seen in color
   by zooming on a computer screen. Artefact reduction compared to CDnCNN can be seen in the top and bottom pictures (see in particular the flower's pistil).}	\label{fig:supp_color}

\end{figure*}

\def \1{g011}
\def \2{g006}
\def \3{g020}
\def \4{g23}

\def \width{2.0}
\def \Width{1.9}

\begin{figure*}[h!]
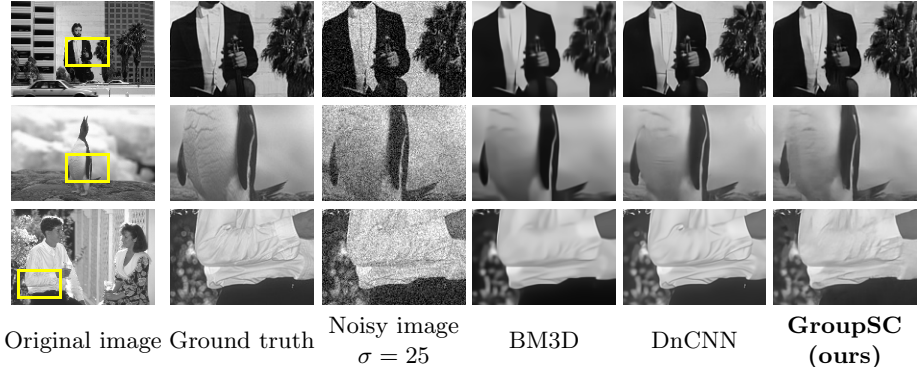

\centering
\begin{tabular}{cccccc}
	\includegraphics[width=\Width cm]{figures/gray_denoising/\1_bb.png}&
	\includegraphics[width=\Width cm]{figures/gray_denoising/\1_gt.png}&
	\includegraphics[width=\Width cm]{figures/gray_denoising/\1_noise.png}&
	\includegraphics[width=\Width cm]{figures/gray_denoising/\1_bm3d.png}&
	\includegraphics[width=\Width cm]{figures/gray_denoising/\1_dn.png}&
	\includegraphics[width=\Width cm]{figures/gray_denoising/\1_group.png}
	\\
	
	\includegraphics[width=\Width cm]{figures/gray_denoising/\2_bb.png}&
	\includegraphics[width=\Width cm]{figures/gray_denoising/\2_gt.png}&
	\includegraphics[width=\Width cm]{figures/gray_denoising/\2_noise.png}&
	\includegraphics[width=\Width cm]{figures/gray_denoising/\2_bm3d.png}&
	\includegraphics[width=\Width cm]{figures/gray_denoising/\2_dn.png}&
	\includegraphics[width=\Width cm]{figures/gray_denoising/\2_group.png}
	\\	
	\includegraphics[width=\Width cm]{figures/gray_denoising/\4_bb.png}&
	\includegraphics[width=\Width cm]{figures/gray_denoising/\4_gt.png}&
	\includegraphics[width=\Width cm]{figures/gray_denoising/\4_noise.png}&
	\includegraphics[width=\Width cm]{figures/gray_denoising/\4_bm3d.png}&
	\includegraphics[width=\Width cm]{figures/gray_denoising/\4_dn.png}&
	\includegraphics[width=\Width cm]{figures/gray_denoising/\4_group.png}
	\\	
    Original image &	
    Ground truth &
   \begin{tabular}{cc}Noisy image\\ $\sigma=25$  \end{tabular}&
   BM3D & 
   DnCNN &
   \begin{tabular}{cc}\textbf{GroupSC}\\ \textbf{(ours)} \end{tabular} \\
\end{tabular}
\vspace*{-0.4cm}

\caption{Grey denoising results for 3 images from the BSD68 dataset. Best seen 
by zooming on a computer screen. GroupSC's images are slightly more detailed than DnCNN on the top and middle image, whereas DnCNN does subjectively slightly better on the bottom one. Overall, these two approaches perform similarly on this dataset.}	\label{fig:supp_gray}

\end{figure*}




\def \1{23}
\def \2{14}
\def \3{07}
\def \4{20}
\def \width{2.0}
\def \Width{1.9}

\def \1{60}
\def \2{70}
\def \3{92}
\def \4{26}
\def \width{2.5}
\def \Width{1.9}

\begin{figure*}[h!]
\centering
\begin{tabular}{cccccc}
	\includegraphics[width=\Width cm]{figures/mosa_app/\1_bb.png}&
	\includegraphics[width=\Width cm]{figures/mosa_app/\1_gt.png}&
	\includegraphics[width=\Width cm]{figures/mosa_app/\1_noise.png}&
	\includegraphics[width=\Width cm]{figures/mosa_app/\1_sc.png}&
	\includegraphics[width=\Width cm]{figures/mosa_app/\1_ircnn.png}&
	\includegraphics[width=\Width cm]{figures/mosa_app/\1_group.png}
	\\
    \includegraphics[width=\Width cm]{figures/mosa_app/\2_bb.png}&
	\includegraphics[width=\Width cm]{figures/mosa_app/\2_gt.png}&
	\includegraphics[width=\Width cm]{figures/mosa_app/\2_noise.png}&
	\includegraphics[width=\Width cm]{figures/mosa_app/\2_sc.png}&
	\includegraphics[width=\Width cm]{figures/mosa_app/\2_ircnn.png}&
	\includegraphics[width=\Width cm]{figures/mosa_app/\2_group.png}
	\\
    \includegraphics[width=\Width cm]{figures/mosa_app/\3_bb.png}&
	\includegraphics[width=\Width cm]{figures/mosa_app/\3_gt.png}&
	\includegraphics[width=\Width cm]{figures/mosa_app/\3_noise.png}&
	\includegraphics[width=\Width cm]{figures/mosa_app/\3_sc.png}&
	\includegraphics[width=\Width cm]{figures/mosa_app/\3_ircnn.png}&
	\includegraphics[width=\Width cm]{figures/mosa_app/\3_group.png}
    \\

    Original image &	
	Ground truth &
	Corrupted&
	SC & 
	IRCNN &
	\begin{tabular}{cc}\textbf{GroupSC}\\ \textbf{(ours)} \end{tabular} \\
\end{tabular}
\vspace*{-0.4cm}

\caption{Color denoising results for 3 images from the Urban100 dataset. Best seen in color
by zooming on a computer screen. On the three images, our approach groupSC exhibits significantly less artefacts than IRCNN and our baseline SC.}	\label{fig:supp_mosa}

\end{figure*}

\section{Parameters visualization}\label{sec:viz}

\begin{figure}[h!]
    \centering
    \begin{tabular}{cc}
        \includegraphics[width=0.3\linewidth]{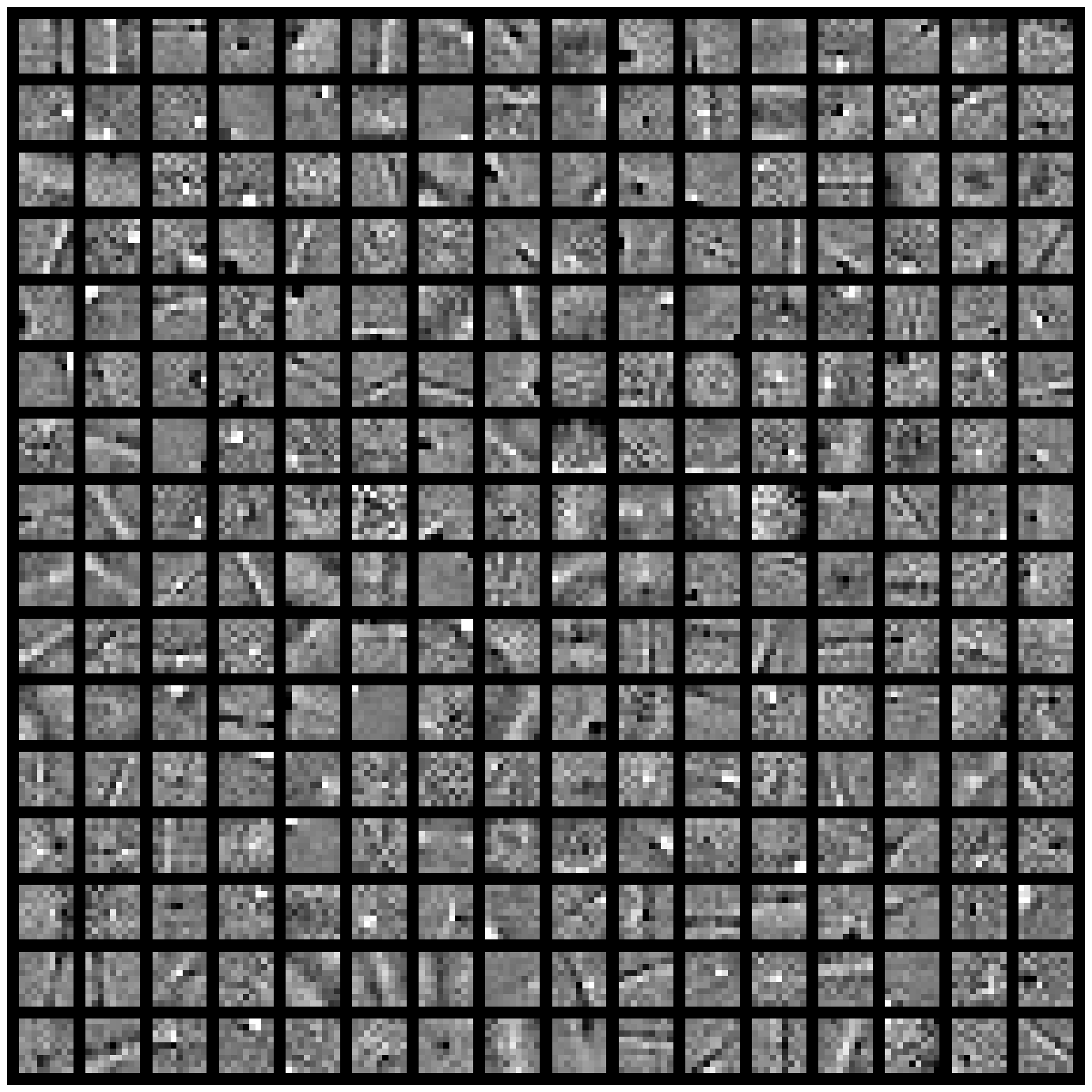} &
        \includegraphics[width=0.3\linewidth]{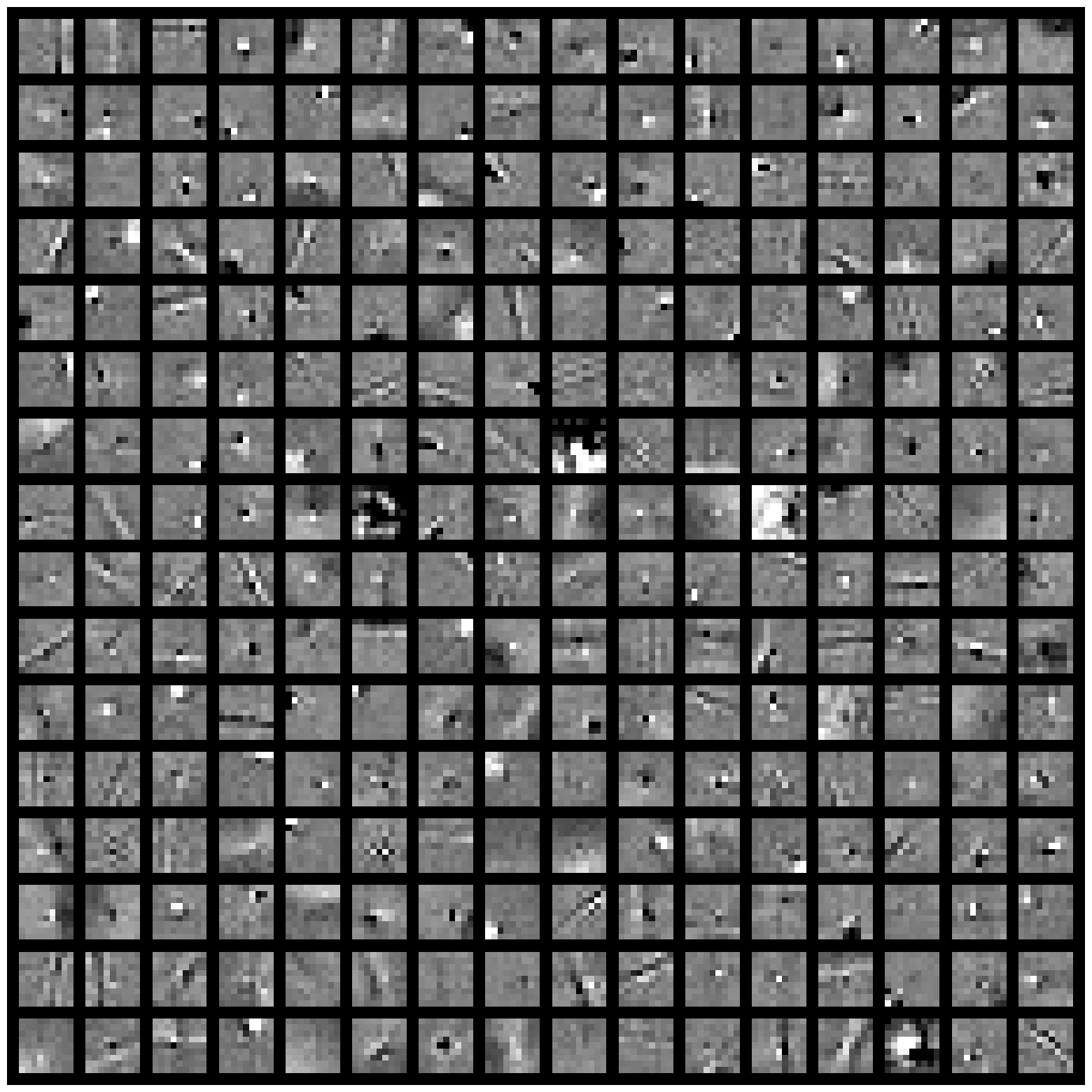}   \\
        $\Db$                                                     &
        $\Wb$                                                       \\
    \end{tabular}
    \caption{Learned dictionnaries of groupSC for denoising.}\label{fig:dico}
\end{figure}

\begin{figure}[h!]
    \centering
    \begin{tabular}{cc}
        \includegraphics[width=0.5\linewidth]{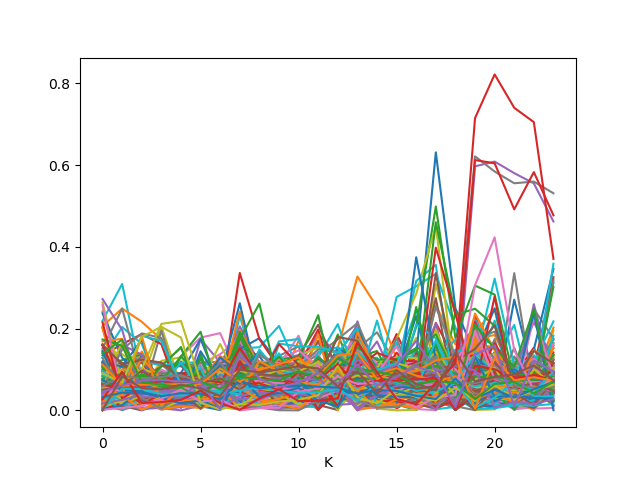} &
        \includegraphics[width=0.5\linewidth]{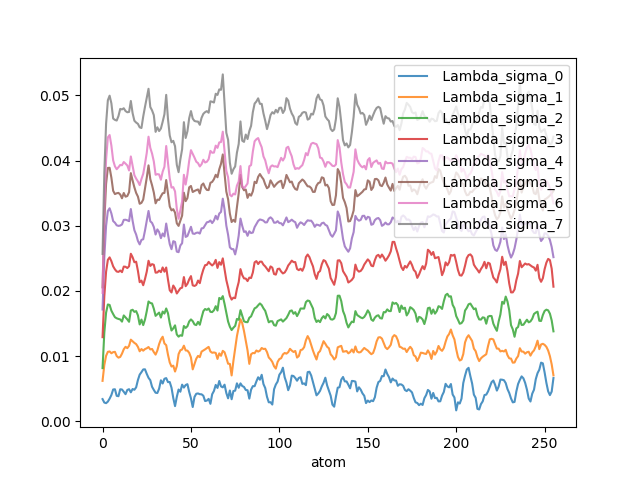}                                                        \\
        \begin{tabular}{c} Sequence of regularization parameters \\
            $\Lambdab_i$ of a non-blind models.\end{tabular}&
        \begin{tabular}{c}  Set of regularization parameters\\$(\Lambda_{\sigma_0}, \dots, \Lambda_{\sigma_n} )$ \\  of a blind model. \end{tabular}
    \end{tabular}
    \caption{Learned regularization parameters of groupSC for denoising and blind denoising. Models are trained on BSD400.}\label{fig:lmbda}
\end{figure}

\noindent
\begin{minipage}{0.70\textwidth}
    We present in this section some visualizations of the learned parameters of our introduced model groupsc for a dnoising task.
    We reported in Figure~\ref{fig:dico} learned dictionaries $\Db$ and $\Wb$ (model trained with $\Cb = \Db$).
    We observe that dictionaries are coupled.
    We reported in Figure~\ref{fig:lmbda} the sequence of regularization parameters $(\Lambdab_{k})_{k=0,1 \dots K-1}$ for a  denoising task, and 
    $(\Lambda_{\sigma_0}, \dots, \Lambda_{\sigma_n} )$.
    for blind denoising. Finally, we reported in Figure~\ref{fig:kappa} the learned weights $\kappab$ of the gaussian kernel for comparing patches.
\end{minipage}
\hfill
\begin{minipage}{0.25\textwidth}

    \includegraphics[width=4cm]{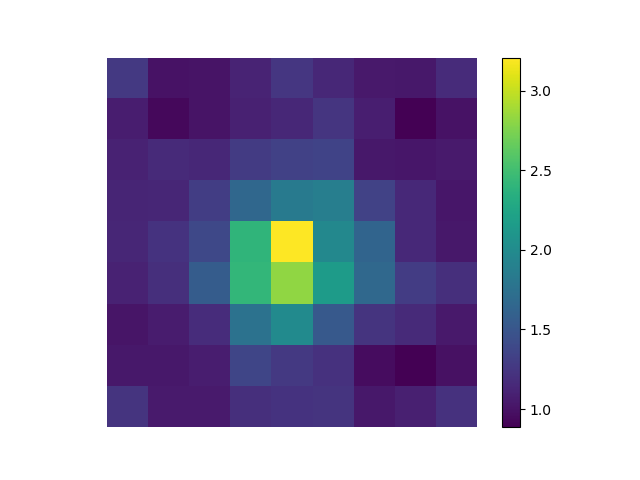}\label{fig:kappa}
    \captionof{figure}{Weights $\kappab$ for comparing patches.}\label{fig:kappa}

\end{minipage}

\end{document}